%% file: paper.tex
\documentclass{article}

\usepackage{arxiv}

\usepackage[utf8]{inputenc} % allow utf-8 input
\usepackage[T1]{fontenc}    % use 8-bit T1 fonts
\usepackage{hyperref}       % hyperlinks
\usepackage{url}            % simple URL typesetting
\usepackage{booktabs}       % professional-quality tables
\usepackage{amsfonts}       % blackboard math symbols
\usepackage{nicefrac}       % compact symbols for 1/2, etc.
\usepackage{microtype}      % microtypography
\usepackage{lipsum}
\usepackage{graphicx}
\usepackage{graphicx}
\usepackage{balance}  % for  \balance command ON LAST PAGE  (only there!)

\usepackage{comment}
\usepackage{tabularx}
\usepackage{paralist}
\usepackage{booktabs}
\usepackage{algorithm, algorithmic}
\usepackage{amssymb}
\usepackage{amsthm}

\newcommand{\eat}[1]{\ignorespaces}

\usepackage{array}
\newcolumntype{P}[1]{>{\centering\arraybackslash}p{#1}}
\newcolumntype{M}[1]{>{\centering\arraybackslash}m{#1}}
\newcolumntype{R}[1]{>{\arraybackslash}m{#1}}

\usepackage{enumerate}

\usepackage{comment}
\usepackage{paralist}
\usepackage{nicefrac}

\usepackage{nicefrac}

\usepackage{mathtools}
\usepackage{blkarray, bigstrut} 

\usepackage{tikz}
\usepackage{verbatim}
\usetikzlibrary{arrows}
\usetikzlibrary{shapes,snakes}
\usetikzlibrary{decorations.pathmorphing} 
\usetikzlibrary{fit}					
\usetikzlibrary{backgrounds}	

\makeatletter
\global\let\tikz@ensure@dollar@catcode=\relax
\makeatother

\usepackage{subfigure}
\usepackage{xcolor}

\definecolor{thelightblue}{RGB}{0,191,255}
\definecolor{theblue}{RGB}{0,0,180}
\definecolor{mygrey}{gray}{0.6}

\usepackage{blkarray}
\usepackage{bigstrut, booktabs}

\makeatletter
\renewcommand*\env@matrix[1][*\c@MaxMatrixCols c]{
\hskip -\arraycolsep
\let\@ifnextchar\new@ifnextchar
\array{#1}}
\makeatother

\usepackage{booktabs} 
\usepackage{enumitem}

\usepackage{subfigure}
\usepackage{rotating}

\usepackage{multirow}

\usepackage{color}
\usepackage{xcolor}
\definecolor{mydarkblue}{RGB}{0, 20, 159} 
\definecolor{mydarkblue}{rgb}{0,0.08,0.45} 
\definecolor{mydarkblue}{rgb}{0,0.08,0.45} 

\usepackage{hyperref}
\hypersetup{%
    colorlinks=true,
    linkcolor=mydarkblue,
    citecolor=mydarkblue,
    filecolor=mydarkblue,
    urlcolor=mydarkblue}

\DeclareSymbolFont{cmbrightop}{OT1}{cmbr}{m}{n}
\DeclareMathSymbol{\sfPsi}{\mathalpha}{cmbrightop}{9}
\let\hat\widehat

\usepackage{tikz}
\usepackage{verbatim}
\usetikzlibrary{arrows}
\usetikzlibrary{shapes,snakes}
\usetikzlibrary{decorations.pathmorphing} 
\usetikzlibrary{fit}					
\usetikzlibrary{backgrounds}	

\definecolor{gray}{RGB}{150,150,150}
\definecolor{theblue}{RGB}{0, 20, 159} 

\definecolor{myyellow}{RGB}{255,255,204}
\definecolor{myred}{RGB}{255,204,204}
\definecolor{myblue}{RGB}{0,200,255}
\definecolor{mygreen}{RGB}{80,220,80}

\theoremstyle{definition}
\newtheorem{definition}{Definition}[section]

\usepackage{balance}

\usepackage{epsfig}

\usepackage{graphicx}

\usepackage{amssymb}
\usepackage{amsmath}
\usepackage{amsfonts}

\usepackage{tabularx}
\usepackage{booktabs}
\usepackage{relsize}
\usepackage{tcolorbox}
\usepackage{numprint} 

\usepackage{setspace}
\usepackage{wrapfig}

\graphicspath{{./}{./graphics/}}
\newcolumntype{H}{>{\setbox0=\hbox\bgroup}c<{\egroup}@{}}

\newcommand{\be}{\begin{equation}}
\newcommand{\ee}{\end{equation}}
\newcommand{\bea}{\begin{eqnarray}}
\newcommand{\eea}{\end{eqnarray}}

\relax

\let\hat\widehat
\let\tilde\widetilde

\usepackage{xpatch}
\usepackage{wrapfig}

\usepackage{nicefrac}

\pdfoutput=1

\renewcommand\footnotemark{}

\begin{document}
% \title{Programmable Graph Learning: End-to-end Framework for Heterogeneous Systems\thanks{Correspondence to: xiaoyao@usc.edu and nesreen.k.ahmed2@intel.com}}

\title{End-to-end Mapping in Heterogeneous Systems using Graph Representation Learning}

%Yao Xiao ( University of Southern California ) < xiaoyao@usc.edu> 
%Guixiang Ma ( Intel Labs ) < guixiang.ma@intel.com> 
%Nesreen Ahmed ( Intel Labs ) < nesreen.k.ahmed@intel.com> 
%Mihai Capotă ( Intel Labs ) < mihai.capota@intel.com> 
%Theodore Willke ( Intel Labs ) < ted.willke@intel.com> 
%Shahin Nazarian ( University of Southern California ) < Shahin.nazarian@usc.edu> 
%Paul Bogdan ( USC ) < pbogdan@usc.edu> 

\author{
 Yao Xiao \\
 USC \\
 Los Angeles, CA, 90089 \\
 \texttt{xiaoyao@usc.edu} \\
 \And 
  Guixiang Ma \\
  Intel Labs\\
  Hillsboro, OR 97124 \\
  \texttt{guixiang.ma@intel.com} \\
  \And 
 Nesreen K. Ahmed\thanks{Correspondence to: nesreen.k.ahmed@intel.com, pbogdan@usc.edu, and xiaoyao@usc.edu} \\
  Intel Labs\\
  Santa Clara, CA 95054 \\
  \texttt{nesreen.k.ahmed2@intel.com} \\
  \And 
 Mihai Capotă \\
  Intel Labs\\
  Hillsboro, OR 97124 \\
  \texttt{mihai.capota@intel.com} \\
  \And 
 Theodore Willke \\
  Intel Labs\\
  Hillsboro, OR 97124 \\
  \texttt{ted.willke@intel.com} \\
  \And 
Shahin Nazarian \\
 USC \\
 Los Angeles, CA, 90089 \\
 \texttt{shahin@usc.edu} \\
 \And 
  Paul Bogdan \\
 USC \\
 Los Angeles, CA, 90089 \\
 \texttt{pbogdan@usc.edu} 
}
\vspace{-10mm}

\maketitle

\input{sec-intro}

\input{sec-related}

\input{sec-framework}

\input{sec-exp}
\input{sec-conclusion}

\bibliographystyle{unsrt}
\bibliography{paper}

\appendix
\input{sec-appendix}

\end{document}

%% file: sec-intro.tex
\section{Introduction}
\label{sec:intro}

%\nesreen{Please add more references and citations in the introduction.}

%\nesreen{Please notice that the intro does not say anything about why we use graphs and GNNs. So add a few sentences on the importance of modeling the dependency structure of software programs which would help automatic compilation in heterogeneous platforms.}

The recent technological advances have significantly contributed to a rapid increase in the algorithmic complexity of various applications, from digital signal processing to autonomous aerial, ground and underwater systems~\cite{krishnan2019air}. In order to control and manage this increased algorithmic complexity, heterogeneous computing systems require intelligent, flexible, and highly efficient programming strategies to provide high performance while minimizing energy costs~\cite{haj2019view,xiao2019self}. However, the current monolithic programming models and task mapping to compute engines do not fully exploit the recent architectural innovations and can exacerbate the load imbalance and communication inefficiencies~\cite{xiao2017load}.

In order to fully utilize the capabilities of hardware platforms, the compilation of parallel programs requires expert heuristics to decide how many threads to spawn and how to schedule them onto heterogeneous computing systems~\cite{cummins2017end}. Due to workload imbalance, synchronization overhead, and resource sharing contention, the overall performance may lead to sub-optimal executions. 
% you may refer to some prior work (SOSPCS or some other work on the "workload imbalance issues, etc. 

% To address these issues, researchers \cite{cummins2017end,cummins2020programl, grewe2013portable} propose and solve the device mapping problem -- where given a software kernel, how to predict the optimal processor, i.e., CPU or GPU, to provide better performance -- by developing machine learning approaches to outperform the inefficient heuristics. 
To address these issues, we need to predict the optimal processor (e.g., CPU or GPU) using machine learning models to provide better performance given a software kernel, which is formally defined as the device mapping problem \cite{mirhoseini2017device}. It is analyzed and solved \cite{cummins2017end,cummins2020programl, grewe2013portable} by using machine learning approaches to outperform the inefficient heuristics. However, as applications become more diverse and complex, it is inefficient to map them only onto one type of processors. For example, autonomous car driving distributes the visualization and recognition tasks, consisting of many \emph{for} loops, onto cores in GPUs to provide higher parallelization. At the same time, sequential decisions based on \emph{if-else} statements require CPUs to provide a fast execution on a single critical thread. There is a tradeoff between GPUs and CPUs.% GPUs provide a higher number of compute engines for parallel computing whereas CPUs have higher frequencies compared to GPUs, leading to a faster execution of sequential threads. 

Therefore, to combine the benefits of both CPUs and GPUs, as opposed to the traditional device mapping problem, we formulate a new problem to be considered within the high performance computing and machine learning contexts: 

\begin{tcolorbox}
Given a complex software application, the goal is to learn a mapping function that predicts which code segments would run best on a specific hardware device in heterogeneous hardware platforms.
\end{tcolorbox}

%\nesreen{Add a reference to multi-fractal analysis.}
Computations in programs can be considered as a graph where each node represents a compute instruction and each edge represents an information flow from one instruction to another. This graph representation of programs enables us to model the dynamic dependency structures of software programs and helps analyze program characteristics and automatically compile programs in heterogeneous platforms. The automation is achieved via graph learning models to predict the type of each program from an initial feature matrix. In order to obtain the representative feature matrix from a graph, we apply multi-fractal analysis \cite{xue2017reliable} to quantitatively measure the topological structures hidden in a graph.

\begin{figure*}
\centering
\includegraphics[width=0.80\textwidth, height=0.36\textwidth]{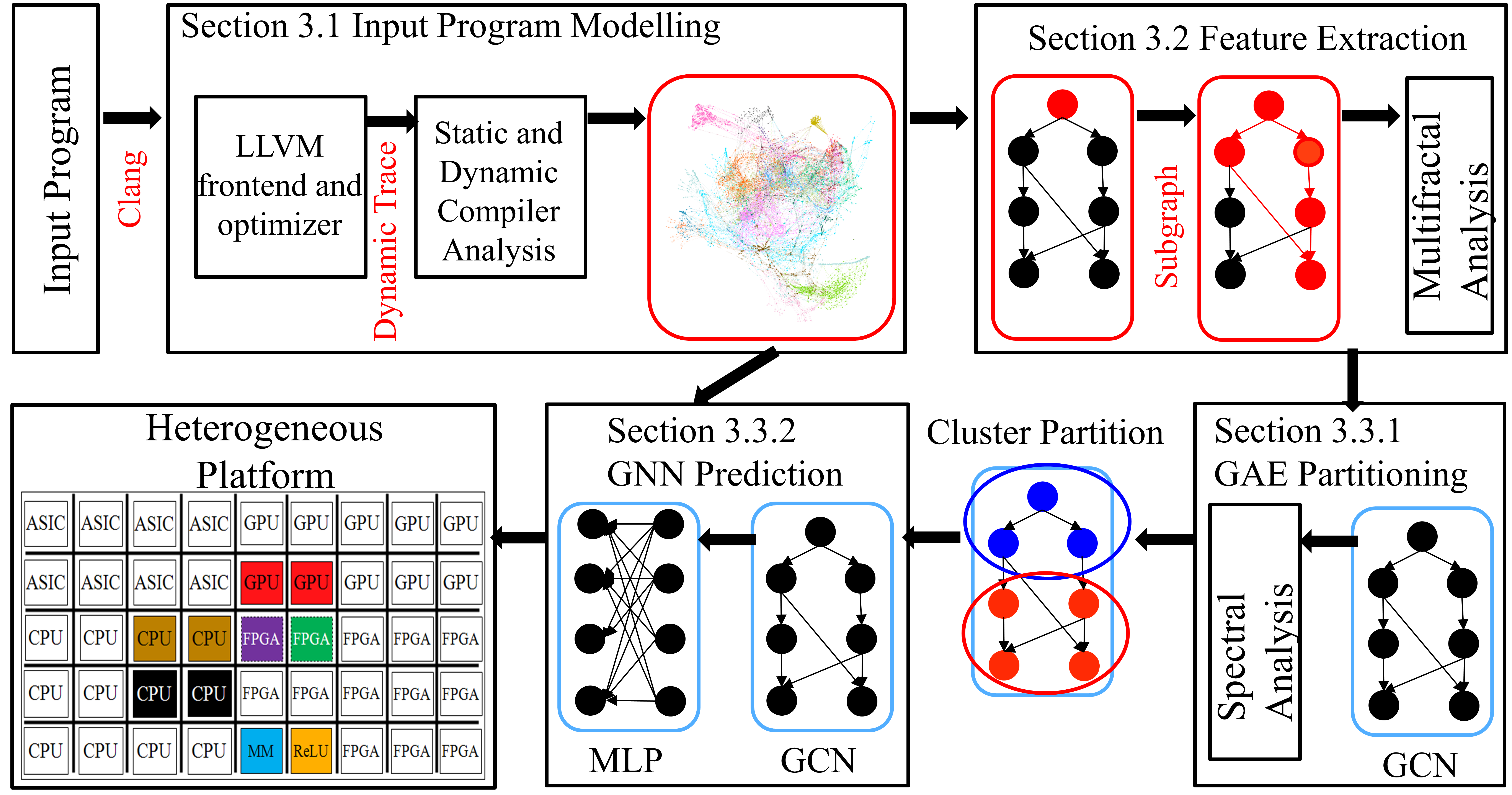}%0.4
\vspace{-2mm}
\caption{Overview of the proposed Programmable Graph Learning framework (PGL). PGL constructs a dynamic dataflow graph for each input software program via LLVM IR. PGL then utilizes a novel feature extraction algorithm based on random walks and multi-fractal analysis to construct node features that capture the topological dependencies and structures in dynamic dataflow graphs. These features are further used by a graph autoencoder to partition the graph into clusters (i.e., software kernels) and a graph neural network model to predict the best hardware device for each kernel.}
% \vskip -10pt
\label{fig-overview}
\vspace{-3mm}
\end{figure*}

To solve this challenging optimization problem, we propose a \emph{unified, end-to-end, programmable graph representation learning} (PGL) framework capable of mining the complexity of high level programs down to the universal IR, extracting the specific computational patterns, and predicting which code segments run best on a specific core in heterogeneous hardware platforms. %We first model each application as a dynamic dataflow graph where nodes represent low level virtual machine (LLVM) intermediate representation (IR) instructions and edges represent control, data, and memory dependencies. 
The proposed PGL framework, shown in Figure 1, is flexible and capable of working with various graph representations of software codes (e.g., regardless of abstract syntax tree, data-control flow graph). We also propose and evaluate a dynamic data flow graph representation constructed from a partially executed trace of a code, where nodes represent LLVM intermediate representation (IR) instructions and edges represent control, data, and memory dependencies, which can better identify the structural information flow and capture memory dependencies.  
% \textcolor{red}{This framework is capable of understanding various graph representations of programs, regardless of abstract syntax tree or data-control flow graph. We also propose and evaluate the proposed graph representation from an partially executed trace of a program to identify the structural information flow.} 

%\textcolor{red}{Nesreen: Could you please refer to the main figure here?}
%In addition, the proposed PGL framework in Figure \ref{fig-overview} utilizes a novel \emph{topological feature} extraction algorithm to capture the local structural patterns in the code graphs. For each node, we perform different random walks scenarios  to explore its local structures. Once we find the destination node, we find the subgraph by performing backtracking to reach to the source node. We then exploit the multi-fractal analysis to determine the generalized fractal dimension of the subgraph. Finally, we build a graph autoencoder to partition the graph from the complex application into several kernels and a graph neural network (GNN) to predict the correct label for each kernel.

We evaluate the proposed PGL framework on a heterogeneous platform consisting of 32 CPUs and 32 GPUs. The GNN is first trained with $C$ styled kernels converted from \emph{OpenCL} from seven benchmark suites to learn the weights of GNNs. Next, we integrate the trained GNN model with the GAE into the framework and test new incoming applications. Experimental results demonstrate a maximum speedup of 6.42x when compared to the thread-based execution and 2.02x higher compared to the state-of-the-art technique.

%\nesreen{I think you are simplifying the problem definition too much, it's not just about GPU and CPU, it's about heterogeneous hardware platforms beyond those two specific examples. Try to make this more clear in the intro and elsewhere}

%\colorbox{green}{Yao: I have already addressed it. Please take a look at the problem formulation and contribution and some sentences in abstract.}

%\nesreen{We propose a framework to solve the challenging task of heterogeneous device mapping. Our proposed framework learns to automatically map the computations of complex software applications to the appropriate hardware device in heterogeneous platforms.}

\vspace{-3mm}
\paragraph{Contributions.} Our main contributions are as follows:
\begin{itemize}[topsep=0pt,itemsep=1ex,partopsep=0pt,parsep=0pt]%[topsep=0pt,itemsep=0pt] 
    \item We formulate a new challenging system optimization problem to be considered in the areas of machine learning and computing systems: \textit{Given} a software program, the \textit{goal} is to learn a mapping function that predicts which code segment should run on which hardware device in a heterogeneous computing system.
    % \item We formulate a new system optimization problem to be considered in the areas of machine learning and computing systems: Given a complex program, predict which portions of the code should run on which core on a heterogeneous hardware platform.
    \item We propose a unified, end-to-end,  programmable graph representation learning framework (PGL) that automatically maps the computations of complex software applications to the appropriate hardware device in heterogeneous hardware platforms.
    %to solve this problem by integrating graph autoencoders and graph neural networks (GNNs).
    
    \item The proposed PGL framework uses a novel topological feature extraction algorithm based on random walks and multi-fractal graph analysis to capture the local topological structures of a graph obtained from a program through advanced static and dynamic compiler analysis techniques.
    
    %\item We conduct extensive experiments and baseline comparisons to validate the PGL framework which achieves an application performance improvement up to $6.42$x when compared to the thread-based execution and $2.02$x compared to the state-of-the-art techniques.
\end{itemize}

%% file: sec-related.tex
\section{Related Work}
\label{sec:related}
We summarize the related work into two areas: (1) deep learning models in compiler optimization, and (2) graph representation learning for code representation.

\paragraph{Deep Learning in Compiler Optimization.} Heuristics used in compilers require expert knowledge to optimize programs on heterogeneous systems and often lead to sub-optimal performance due to synchronization overhead and resource management. Machine learning techniques, in particular, deep learning methods, are being applied during the optimization phase to generate efficient machine code \cite{ashouri2018survey,li2020deep,haj2019view}. %The recent work in~\cite{vasilache2018tensor} introduced a high-level language to describe tensor computations used in machine learning and developed an end-to-end compilation flow with an auto-tuning framework to convert a mathematical description of deep learning graphs into a CUDA kernel with memory management and synchronization. More recently, 
The recent work in~\cite{zhou2020transferable} proposed an end-to-end deep reinforcement learning (DRL) method for ML compiler graph optimizations where the learned policies are generalized to new graphs and transferable to different tasks. \cite{haj2020neurovectorizer,haj2019learning} proposed an end-to-end framework  utilizing DRL for handling loop vectorization. In addition, machine learning techniques are also used to optimize the execution time of tensor computation graphs~\cite{jinnai2019knossos} as well as deep neural networks in TASO \cite{jia2019taso} and SOAP \cite{jia2018beyond}. 

\paragraph{Graph Representation Learning for Code Representation.} While many prior works have employed machine learning methods from natural language processing to represent programs as sequence of lexical tokens \cite{nguyen2018deep,cummins2017end}, recently there emerged a number of graph-based machine learning works that aims to capture the structure of programs along with the syntactic and semantic information in the graph representation \cite{alon2019code2vec,ben2018neural,brauckmann2020compiler}. It has been observed that the graph-based representation learning strategies tend to have superior learning ability on the programs for many code analysis tasks, such as code similarity learning \cite{li2019graph}, program classification \cite{mou2016convolutional}, etc. For instance, \cite{brauckmann2020compiler} uses abstract syntax trees (ASTs) and control-data flow graphs (CDFGs) independently to represent programs and apply GNNs for learning predictive compiler tasks on these graphs, which outperforms the recurrent neural networks (RNNs) on the token sequence representation of the programs. \cite{cummins2020programl} models the program's control, data and call dependencies as a graph, and applies a GNN to learn representations from the graph for both node-level and graph-level tasks including compiler analysis, program classification and device mapping.% It has shown that their representation of the program leads to better performance in those tasks compared to the previous methods.  

However, compared to previous frameworks, we propose a \emph{unified end-to-end programmable graph representation learning} (PGL) framework to solve the challenging task of heterogeneous device mapping. The framework has the offline training and online inference, yet it is fully autonomous, it does not require programmer intervention. Therefore, for a new platform with untrained machine learning models, we first learn the optimal mapping of code segments by adjusting the network weights. Once models are well trained, a new incoming program has to be compiled and executed once, next driven into GAE and GNN to predict which code segments are suitable for CPUs or GPUs. 

%However, compared to previous frameworks, we propose the new heterogeneous device mapping problem that states that given an application, the goal is to learn a mapping function that predicts which code segments benefit most on a specific core in heterogeneous hardware platform. We propose a \emph{unified end-to-end programmable graph representation learning} (PGL) framework to solve the challenging task of heterogeneous device mapping to learn to automatically map the computations of complex software applications to the appropriate hardware device in heterogeneous platforms. 

%The framework has the offline training and online inference, yet it is fully autonomous, it does not require programmer intervention. Therefore, for a new platform with untrained machine learning model, the first thing is to learn the optimal mapping of code segments by adjusting the network weights.
%\textcolor{red}{it seems this sentence is unfinished}
%Once models are well trained, a new incoming program has to be compiled and executed once, next fed into GAE and GNN to predict which code segments are suitable for CPUs or GPUs. This framework is easy for programmers to use because it does not ask them to modify the original program in order to be constructed.  

%% file: sec-framework.tex
\section{Programmable Graph-based Learning Framework (PGL)}
\label{sec:framework}
%\nesreen{Add a few sentences to summarize the overall framework, that will be detailed in the next subsections, and refer to the overview figure (figure 1)}
In this section, we describe the proposed PGL framework, which consists of four steps as shown in Figure~\ref{fig-overview}. The descriptions of these steps are detailed in the next sections. Section~\ref{sub-prog} discusses the general approach to transform an application into a dynamic dataflow graph. Then, in Section~\ref{sub-feat}, we develop a novel node feature extraction algorithm based on random walks and multi-fractal graph analysis to quantitatively measure the local fractal structures of a graph. Sections \ref{sub-part} and \ref{sub-pred} discuss the GAE graph partitioning and GNN heterogeneous device mapping prediction, respectively.

\subsection{Input Program Modelling}
\label{sub-prog}
Recently, various graph representations were proposed to represent and capture the latent information flow in a program (e.g., abstract syntax tree (AST) \cite{alon2019code2vec}, contextual flow graph (XFG) \cite{ben2018neural}, and control and data flow graph (CDFG) \cite{brauckmann2020compiler}). These graph representations allow the compiler to analyze the effectiveness and correctness of programs, as well as enable parallel programming via graph partitioning in high performance computing~\cite{xiao2017load}. However, these statically compiled graphs have several limitations. First, memory dependencies are difficult to be identified. If not handled properly, this can exacerbate the data communication overhead and reduce the application performance. Second, the number of iterations in \emph{for} and \emph{while} loops cannot be statically determined. This plays a significant role in predicting whether the code is running in either CPU or GPU based on the workload. For example, if the number of iterations is small, it is ideal to run the code in CPU, because of the faster clock frequency. Otherwise, GPU is preferred because the number of cores on each chip is much denser to provide higher parallelism. Therefore, in order to overcome these drawbacks, we use information generated from static compiler analysis and dynamic compilation to model the information flow in high-level programs as a dynamic dataflow graph. Next, we propose the following representation. 

\begin{definition}[\textsc{Dynamic Dataflow Graph}]
A dynamic dataflow graph is a weighted directed acyclic graph $G=(V,E,W)$, where each node $v$, associated with an attribute $va$ indicating the type of the node (e.g., add, sub, store, or load), $(v, va) \in V$ represents an LLVM IR instruction; each edge $e$, associated with an attribute $ea$ indicating the type of dependencies (e.g., control, data, or memory), $(e, ea) \in E$ represents a dependency between two instructions; a weight $w \in W$ on each edge $e$ represents the amount of data communication between two instructions and the time to execute the instruction. It allows us to quantify communication overhead in the memory hierarchy with L1, L2, and L3 caches. 
\end{definition}\label{def:dynamic_graph}

\begin{wrapfigure}{r}{0.5\textwidth}
\centering
\includegraphics[width=0.44\textwidth, height=0.28\textwidth]{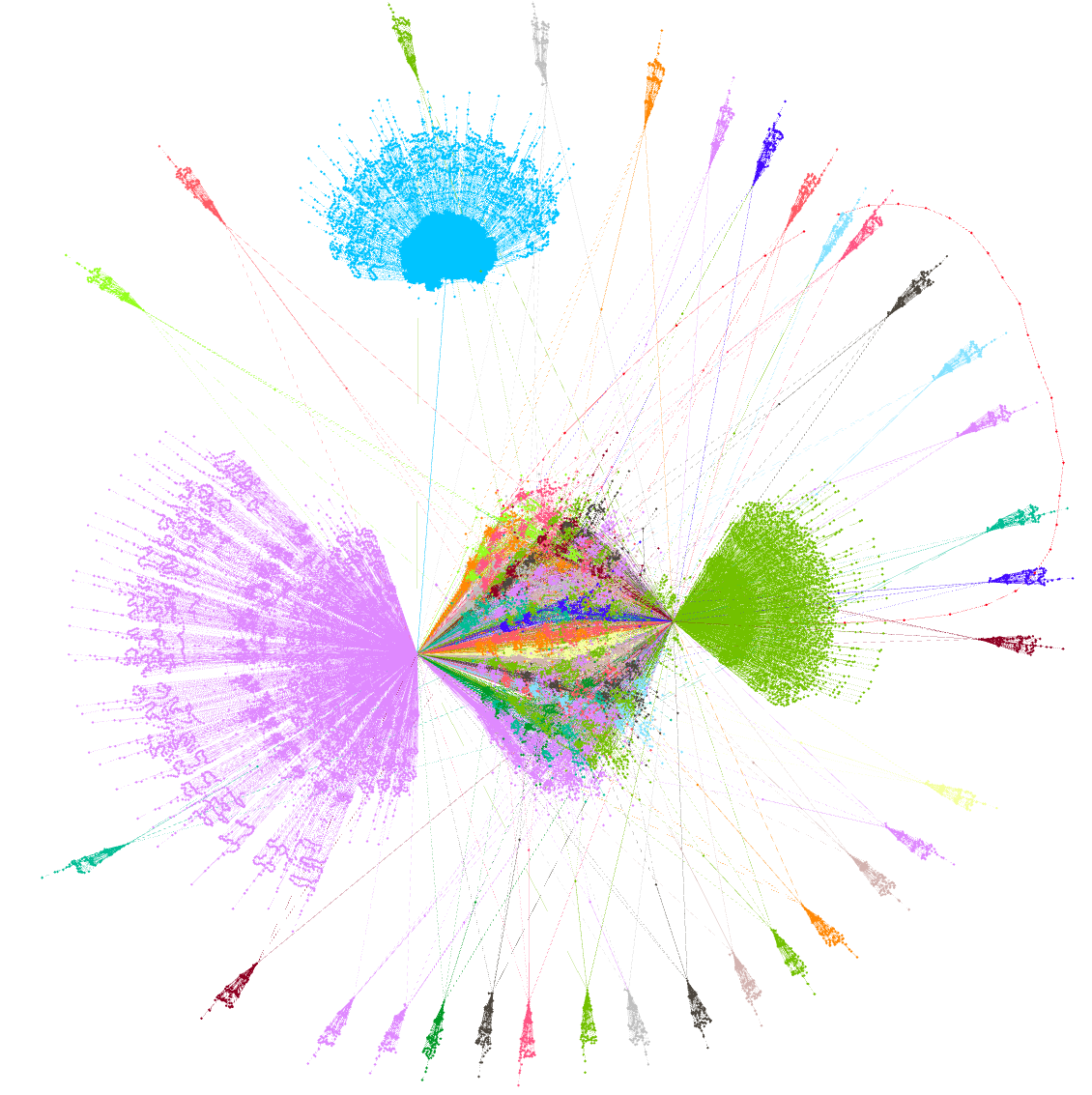}%0.4
\vspace{-3mm}
\caption{An example of a dynamic dataflow graph from a neural network with one hidden layer. Each node represents an LLVM IR instruction and each edge represents a dependency between two instructions. %By adopting this graph representation, we can see that some patterns are recurring due to $for$-loops used in the code. %\textcolor{red}{what's the application that represents the code graph?}
}
% \vskip -10pt
\label{fig-graphex}
\vspace{-3mm}
\end{wrapfigure}

To construct these dynamic dataflow graphs, we first collect the representative dynamic trace generated from executing a program. This trace contains a sequence of LLVM IR instructions to be executed. Then, for each instruction, we check if one of the following dependencies exists and insert a directed edge to construct the graph:
\begin{itemize}[topsep=0pt,itemsep=1ex,partopsep=0pt,parsep=0pt]
    \item \textit{Data dependency}: Source registers of the current instruction depend on the destination registers of the previous instructions.
    \item \textit{Control dependency}: Source registers of the function calls and branches depend on the destination register of the previous instructions.
    \item \textit{Memory dependency}: Memory locations of current store-load instruction are the same as the previous store-load instructions. We perform this memory alias analysis using "\emph{-basicaa -aa-eval -print-allalias-modref-info}" in the LLVM environment.
\end{itemize}

Figure \ref{fig-graphex} shows the graph representation of forward propagation in a neural network with one hidden layer. Note that a node is an LLVM IR instruction, not an operand or a high level language (e.g., C/C++, Java) statement. Different from AST, XFG, and CDFGs, this specific graph representation in Figure \ref{fig-graphex} makes explicit some hidden program information flows from the execution trace generated at run-time and analyzed via data, control, and memory dependencies. One most recurring pattern is the cone structure due to the LLVM IR "\emph{getelementptr}" generated from the pointers in $for$-loops to distribute data to different iterations. 

\subsection{Feature Extraction}
\label{sub-feat}
\indent Each node in a GNN is associated with numerous features, which are further used for clustering or classification to make decisions at node level or graph level. In the literature, the \emph{code2vec} \cite{alon2019code2vec} and \emph{inst2vec} \cite{ben2018neural} are commonly used to extract features by encoding programs via AST paths. However, the trained representations can put larger weights on names rather than code structure, which can lead to misclassification. 

In order to exploit the graph structural information flow of programs, random walks reason about the number of adjacent nodes and the density of connections around a node \cite{grover2016node2vec}. A random walk is defined as a series of nodes, starting from $n_0$, the $j$th node is generated by the following distribution with a fixed length $l$.
\begin{equation}
  P(n_j=j|n_i=i) =
    \begin{cases}
      \frac{w_{ij}}{\sum_{j} w_{ij}} & \text{if $(i,j)\in E$}\\
      0 & \text{otherwise}
    \end{cases}       
\end{equation}
where $w_{ij}$ is the edge weight between node $i$ and node $j$. In addition, multifractal analysis mathematically studies the structural complexity and topological heterogeneity of graphs \cite{xue2017reliable}. The multifractal properties such as generalized fractal dimensions provides the higher order statistics of a graph, which can be quantified by a finite box-covering method. That is, to study the different fractal structures in a graph, the box-covering method uses the box of the same size to cover the graph and then studies the relationship of the size of a box ($l$) and the number of nodes in the $i$th box of size $l$ ($N_i(l)$) as 
\begin{equation}
\sum_i N_i(l)^q \sim l^{\tau(q)}
\end{equation}
where $q$ is the distortion factor to differentiate the topological difference of fractal structures, and $\tau(q)$ is the mass exponent. Next, we can obtain the generalized fractal dimensions $D(q)$ from $\tau(q)$, which characterizes the different fractal structures of a graph.
\begin{equation}
D(q) = \frac{\tau(q)}{q-1}
\end{equation}
%\indent Discuss random walks and multifractal analysis on graphs. \\
\indent Therefore, to mine the local and scale dependent topological properties of programs, we propose Algorithm 1 which exploits random walks and multifractal concepts for encoding topological interdependencies (See the Appendix A for the full details of Algorithm 1.). Random walks explore the local topological density around a node $i$ in a graph by finding random paths starting from node $i$ to node $j$. Once a random path is identified, we backtrack to the final destination node $j$ to find the subgraph $SG$ starting from $i$ to $j$. Next, we perform a multifractal analysis on the subgraph $SG$ to estimate its generalized fractal dimension. %Figure X illustrates this node-based feature extraction strategy. 
The time complexity of Algorithm 1 is bounded by the Dijkstra strategy to find the shortest path for each node to every other nodes, which is $O(ElogV)$, where $E$ and $V$ are the number of edges and nodes, respectively. Finding all shortest paths in a graph has a time complexity of $O(EVlogV)$.

%\nesreen{I'm so confused about those "repeats" in the algorithm (Algorithm 1), it's hard to read.}

\subsection{Graph Representation Learning}

\input{subsec-model}

%% file: subsec-model.tex
Once we extracted the initial node features from the dynamic dataflow graph, we design a deep graph representation learning module with GNNs \cite{wu2020comprehensive} for the graph partition and device mapping prediction problem. Specifically, we propose to use a graph autoencoder (GAE) for partitioning the graph into kernels and a GNN %such as graph convolutional network (GCN) 
to predict the correct label.

% \paragraph{Graph Neural Networks.} GNNs have emerged as a powerful approach for representation learning on graphs in many tasks, such as node classification, link prediction, and graph classification \cite{wu2020comprehensive}. Based on the model architectures and training strategies, existing GNNs can be grouped into different categories, for instance, graph recurrent nerual networks (Graph RNNs), GCNs and GAEs, etc. Graph RNNs capture recursive and sequential patterns of graphs by modeling states at node-level or graph-level \cite{li2015gated}. GCNs define convolution and readout operations on graphs to capture structural features \cite{kipf2016semi}. GAEs learn latent node representations through reconstructing graphs \cite{vincent2008extracting}. In this paper, we take advantage of the learning power of GNNs and propose our representation learning module based upon GNNs for graph partitioning on the LLVM dynamic dataflow graph and solve the device mapping prediction.  

\subsubsection{GAE-based Graph Partitioning}
\label{sub-part}
Graph auto-encoders (GAEs) \cite{zhou2018graph} are a category of GNNs that aims at representing nodes into low-dimensional vectors in an unsupervised training fashion. They are different from other GNNs that are typically used for supervised or semi-supervised learning tasks. In our framework, the goal of the graph partitioning stage is to obtain a good partition for each LLVM graph based on a learned representation that captures the intrinsic structural information of the graph, such that the subgraphs preserve the inherent characteristics of the data, control and memory dependencies in the LLVM graph. To this end, we propose a graph partitioning strategy based on the GAE \cite{kipf2016variational} and spectral clustering \cite{von2007tutorial} for our task, as shown in Appendix B. 

% VGAE: 
% define the learning model as follows:
% \paragraph{Inference Model.} We use the inference model parameterized by a two-layer GCN:
% \begin{equation}
%     q(\mathbf{Z}\vert\mathbf{X,A}) = \prod_{i=1}^{N}q(\mathbf{z}_i\vert\mathbf{X,A})
% \end{equation}
% where $q(\mathbf{z}_i\vert\mathbf{X,A})=\mathcal{N}(\mathbf{z}_i\vert\mathbf{u}_i,diag(\mathbf{\sigma}_i^2))$, $\mathbf{\mu}=\text{GCN}_{\mathbf{\mu}}(\mathbf{X,A})$ is the matrix of mean vectors ${\mu}_i$ and $\text{log}_{\mathbf{\sigma}}=\text{GCN}_{\mathbf{\sigma}}(\mathbf{X,A})$. The two-layer GCN is defined as $\text{GCN}(\mathbf{X,A})=\mathbf{A}'\text{ReLU}(\mathbf{A}'\mathbf{X}\mathbf{W}_0)\mathbf{W}_1$, where $\mathbf{W}_i$ are the weight matrices to be learned and $\text{GCN}_{\mathbf{\mu}}(\mathbf{X,A})$ and $\text{GCN}_{\mathbf{\sigma}}(\mathbf{X,A})$ share first-layer parameters $\mathbf{W}_0$. $\mathbf{A}'=\mathbf{D}^{-\frac{1}{2}}\mathbf{A}\mathbf{D}^{-\frac{1}{2}}$is the symmetrically normalized adjacency matrix and $\text{ReLU}(\cdot)=max(0,\cdot)$.
% \paragraph{Generative Model.} 
% We use the inner product for the generative model and define it as:
% \begin{equation}
%     p(\mathbf{A}\vert\mathbf{Z})=\prod_{i=1}^N\prod_{j=1}^Np(A_{ij}\vert\mathbf{z}_i,\mathbf{z}_j)
% \end{equation}
% where $p(A_{ij}=1\vert\mathbf{z}_i,\mathbf{z}_j)=\sigma(\mathbf{z}_i^\top\mathbf{z}_j)$, $A_{ij}$ are the elements of $\mathbf{A}$ and we use the logistic sigmoid function for $\sigma(\cdot)$.

\subsubsection{GNN-based Device Mapping Prediction}
\label{sub-pred}
% Once the graph is partitioned into different clusters/kernels, next for each kernel, we use graph convolutional network to predict the correct platform to to execute by updating the node vectors iteratively in a fashion of the message passing. At each step, a new node feature vector is generated as a function of its previous vector and the vectors of its neighboring nodes. After a number of epoches to repeat this update of node features, a readout function is used to aggregate the node representations to a single graph-level vector. This graph-level vector is fed through another feed forward neural network to predict the correct label.

Once the graph is partitioned into different clusters/kernels, next for each kernel, we use a GNN to predict the correct platform to execute the kernel by updating the node vectors iteratively in a similar fashion to the message passing. Note that our proposed PGL is a general framework that can leverage various GNN models for the device mapping prediction stage, whereas in this paper, we adopt three different variants of the GNN models: GCN \cite{kipf2016semi}, graph attention network (GAT) \cite{velivckovic2017graph,lee2019attention} and gated graph neural network (GGNN) \cite{li2015gated}, respectively, for this task discussed briefly in Appendix B. We also empirically investigate the comparative effectiveness of these GNN strategies in representation learning on the partitioned LLVM graphs for the graph classification task in heterogeneous device mapping.

%% file: sec-exp.tex
\section{Experiments}
\label{sec:exp}

%\nesreen{In some cases we use ins2vec and other cases we use NCC, should we be more consistent, since both refer to the same fraemwork and paper?}

%\textcolor{green}{Yao: NCC is the overall framework whereas inst2vec is just an embedding used in NCC. I think NCC includes inst2vec and they are not the same.}

\paragraph{Setup.} Given a software program, our goal is to identify the \emph{subgraphs} (i.e., 
code segments) that are optimal to run on CPUs or GPUs.\footnote{Performance varies by use, configuration and other factors. Learn more at \url{www.Intel.com/PerformanceIndex}.} Our developed end-to-end framework discussed in the previous section consists of two components: a GAE and a GNN. Unsupervised learning model GAE is used to partition the new complicated program into several clusters / kernels to be mapped onto heterogeneous systems. Supervised learning model GNN predicts the correct label for each kernel. In the implementation, we use kernels written in OpenCL~\cite{cummins2017end} as training and testing data with 5-fold cross validation for the GNN model. The ground-truth labels are either CPU or GPU for the kernels. In order to evaluate the PGL framework, we first use the GAE model to partition the graphs, to find kernels suitable for either CPU or GPU. Next, different GNN models are used to predict the correct label to the underlying hardware. Table \ref{exp-conf} (in Appendix C.1) lists the configuration parameters of the heterogeneous system used in this section. %In the evaluation, we first report the total accuracy of the GNN on the testing data and then we use programs shown in Table \ref{table-data} (in Appendix C.2) and map kernels onto the heterogeneous system to measure the performance improvement over the state-of-the-art methodologies. We report the normalized speedup in terms of the application performance as the thread-based execution time (slowest) divided by the execution time for each approach.
\paragraph{Datasets.}
We start by using the $256$ heterogeneous device mapping OpenCL kernels in~\cite{cummins2017end} for training and validation of GNNs. These kernels are labelled with CPU vs. GPU; we use the NVIDIA dataset. We then manually convert these kernels to C code. Furthermore, we use standard application benchmarks in Table \ref{table-data} (in Appendix C.2) to validate the overall PGL framework.
% We manually convert to C code 256 OpenCL kernels labelled with CPU and GPU from seven benchmark suites \cite{cummins2017end} as the training and testing input programs. We use the NVIDIA set with an Intel Core i7-3820 CPU and an NVIDIA GTX 970 GPU. Furthermore, we use standard application benchmarks (see Table \ref{table-data} for details) to validate the overall proposed PGL framework in comparison with baselines.

\paragraph{Baseline Comparisons.}

%\nesreen{I believe Graph Attention Network is called GAT instead of GAT}

When comparing the accuracy of the prediction results from GNN models, we use the following: (1) GCN; (2) GAT; and (3) GGNN. We compare our graph representation to the ProGraML \cite{cummins2020programl} graph representation, NCC \cite{ben2018neural}, and DeepTune \cite{cummins2017end}, state-of-the-art techniques to represent programs as graphs. To quantify the benefits of graph partitioning, we compare the PGL framework with the following baselines in terms of the application performance: (1) K-means clustering connected with GCNs (KM+GCN); (2) hierarchical divisive clustering where all observations start in one cluster, and divisions are performed recursively as one moves down the hierarchy, connected with GCNs (HDC+GCN); (3) modularity-based community detection where an optimization model is proposed to measure the structure of graphs \cite{fortunato2010community,xiao2017load}, connected with GCNs (MOD+GCN); (4) METIS graph partitioning \cite{lasalle2015improving} connected with GCNs (METIS+GCN); (5) feed-forward neural network, connected with GCNs \cite{xiao2019self} (NN+GCN). In addition, we compare the PGL framework in terms of the application performance with the following baselines: (1) threads in parallel programming (PAR); (2) modularity based community detection to partition the graph into clusters and a heuristic mapping \cite{xiao2017load} (CommDet); (3) sliding window based neural network to locate specialized structures with a reinforcement learning based mapping (NN+RL) \cite{xiao2019self}; (4) gem5-aladdin, an end-to-end SoC simulation \cite{shao2016co}. 

\renewcommand{\floatpagefraction}{.9}%
\begin{figure*} [!t]
  \centering
  \mbox{
    \includegraphics[width=0.31\textwidth, height=0.31\textwidth]{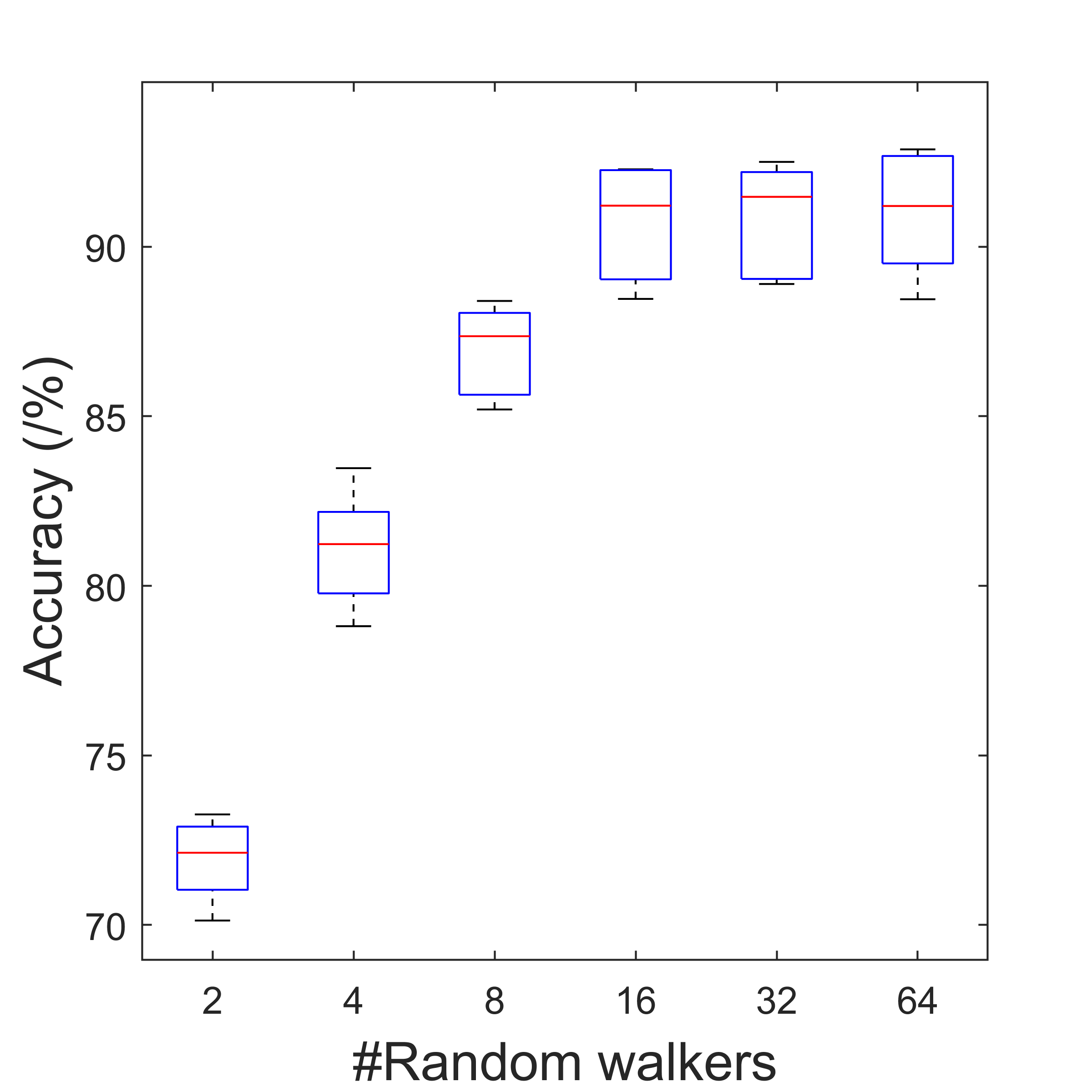}
    \includegraphics[width=0.31\textwidth, 
    height=0.31\textwidth]{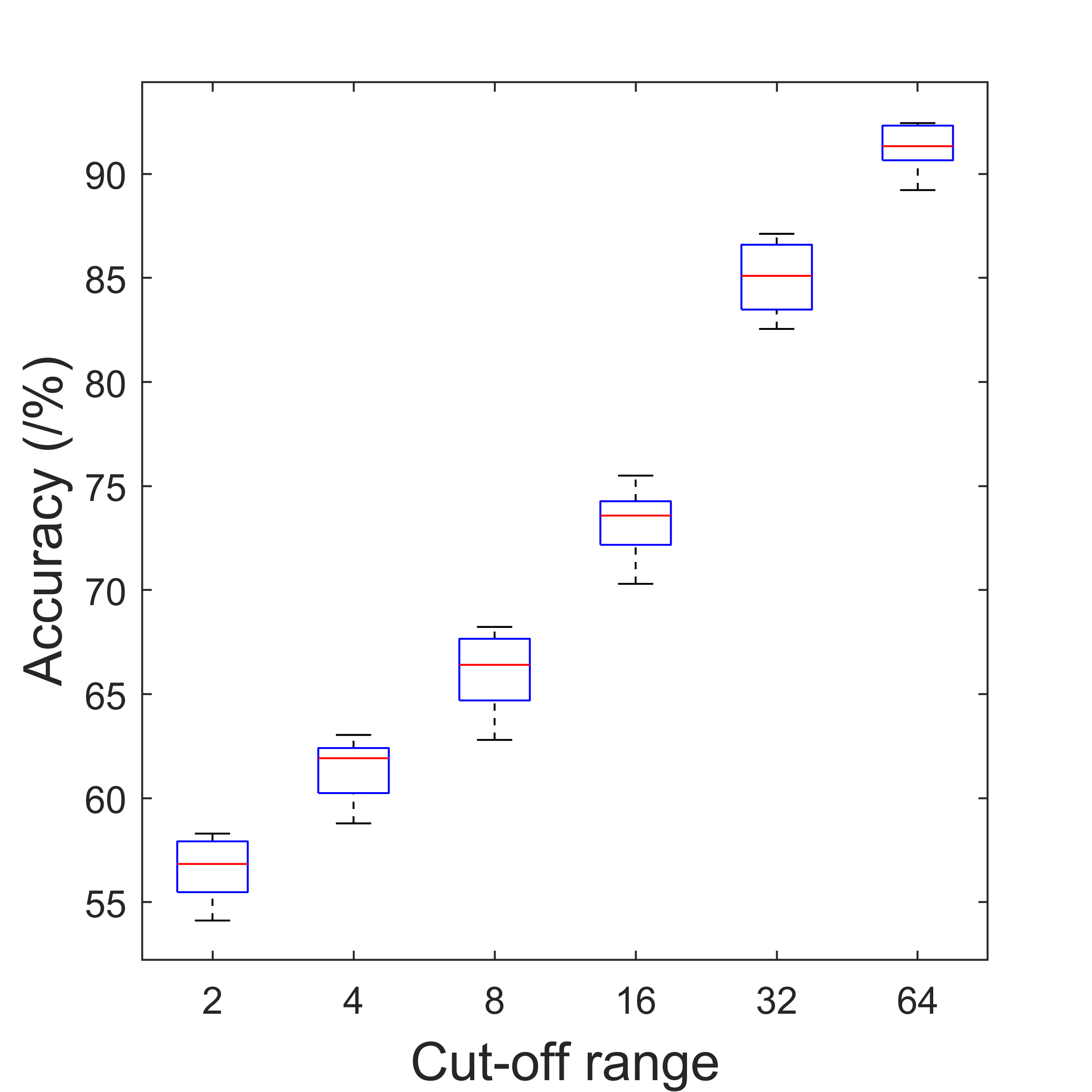}
    \includegraphics[width=0.31\textwidth, height=0.31\textwidth]{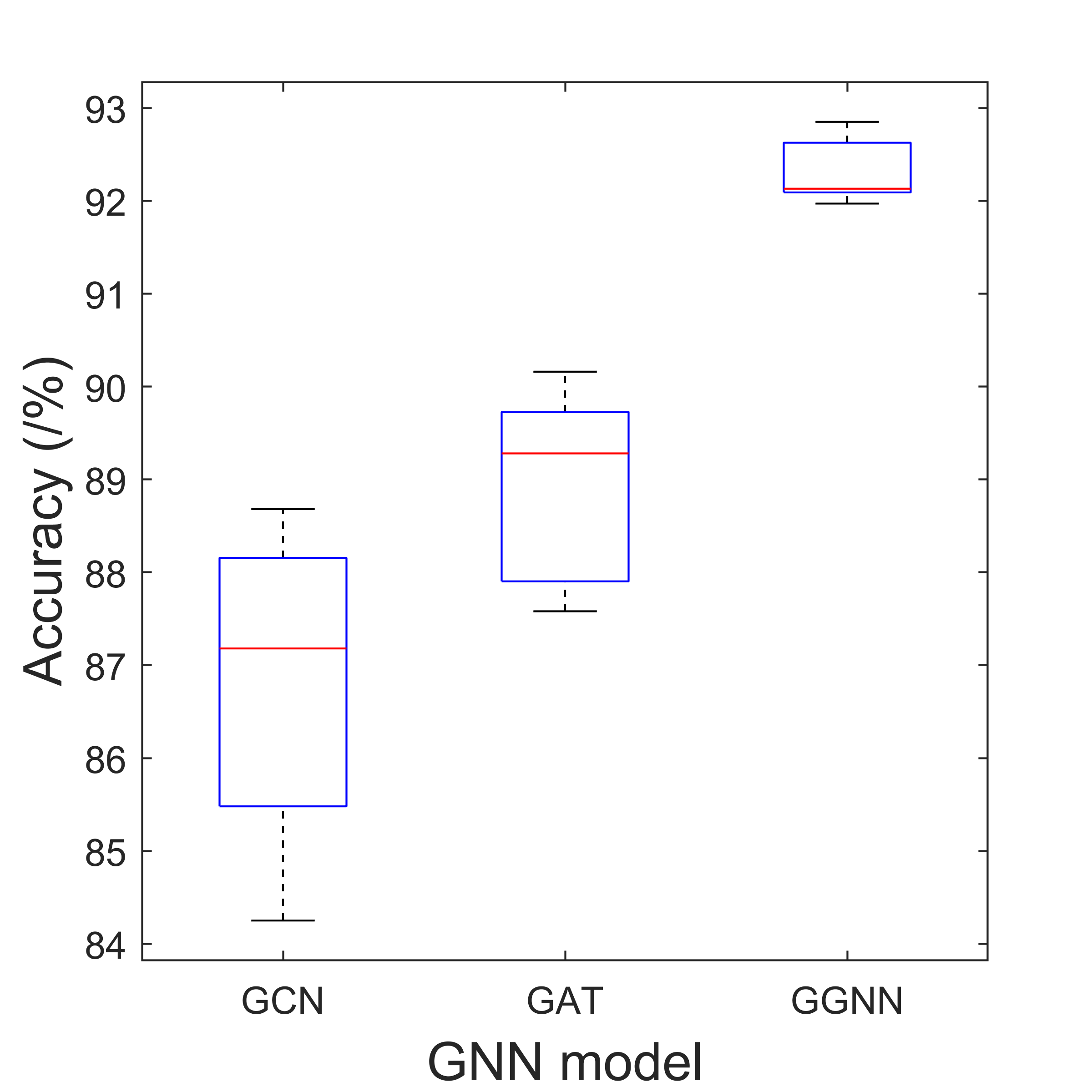}\quad
  }
  \vskip -2pt
  \mbox{
    \includegraphics[width=0.31\textwidth, height=0.31\textwidth]{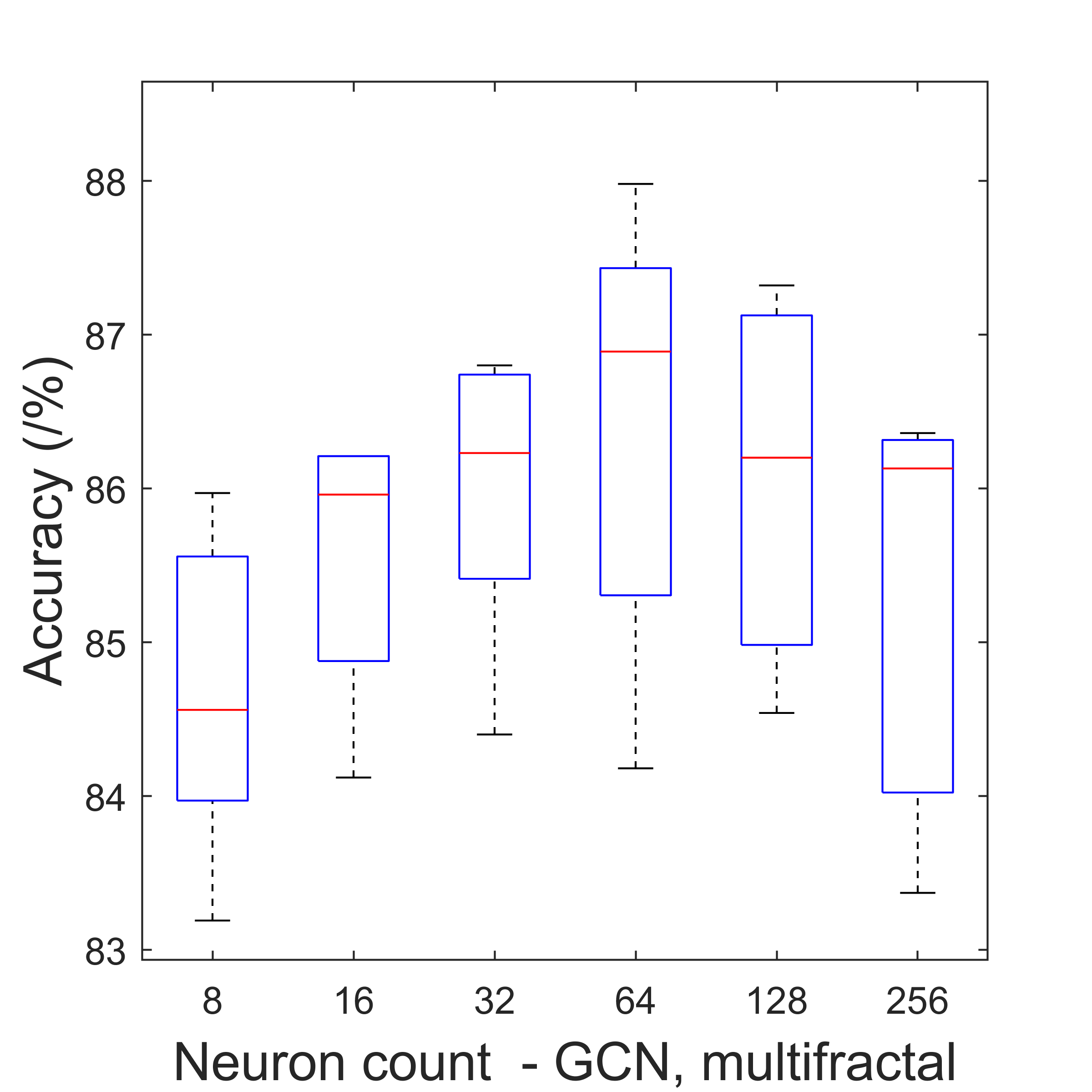}%GCN
    \includegraphics[width=0.31\textwidth,  height=0.31\textwidth]{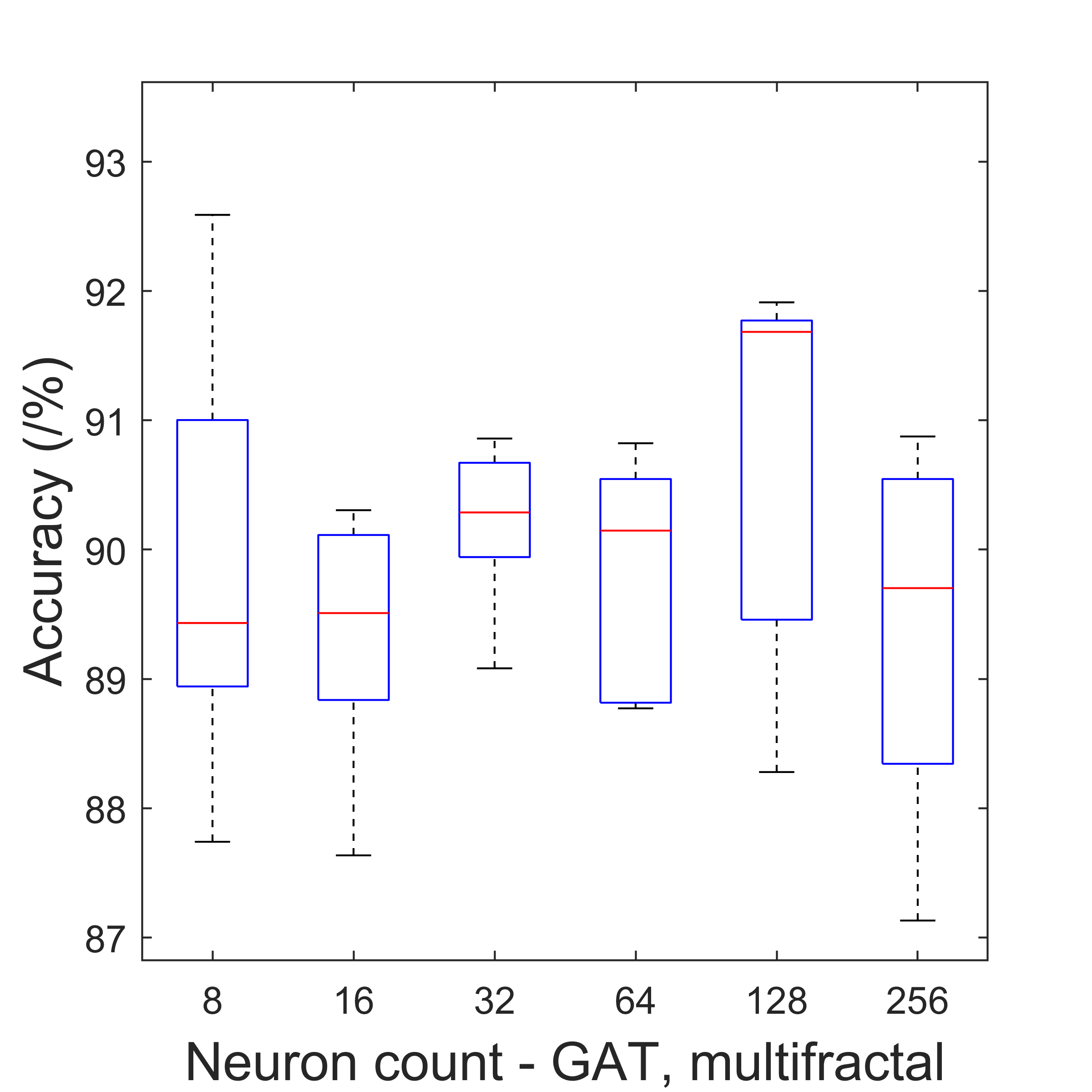}%GAT
    \includegraphics[width=0.31\textwidth,  height=0.31\textwidth]{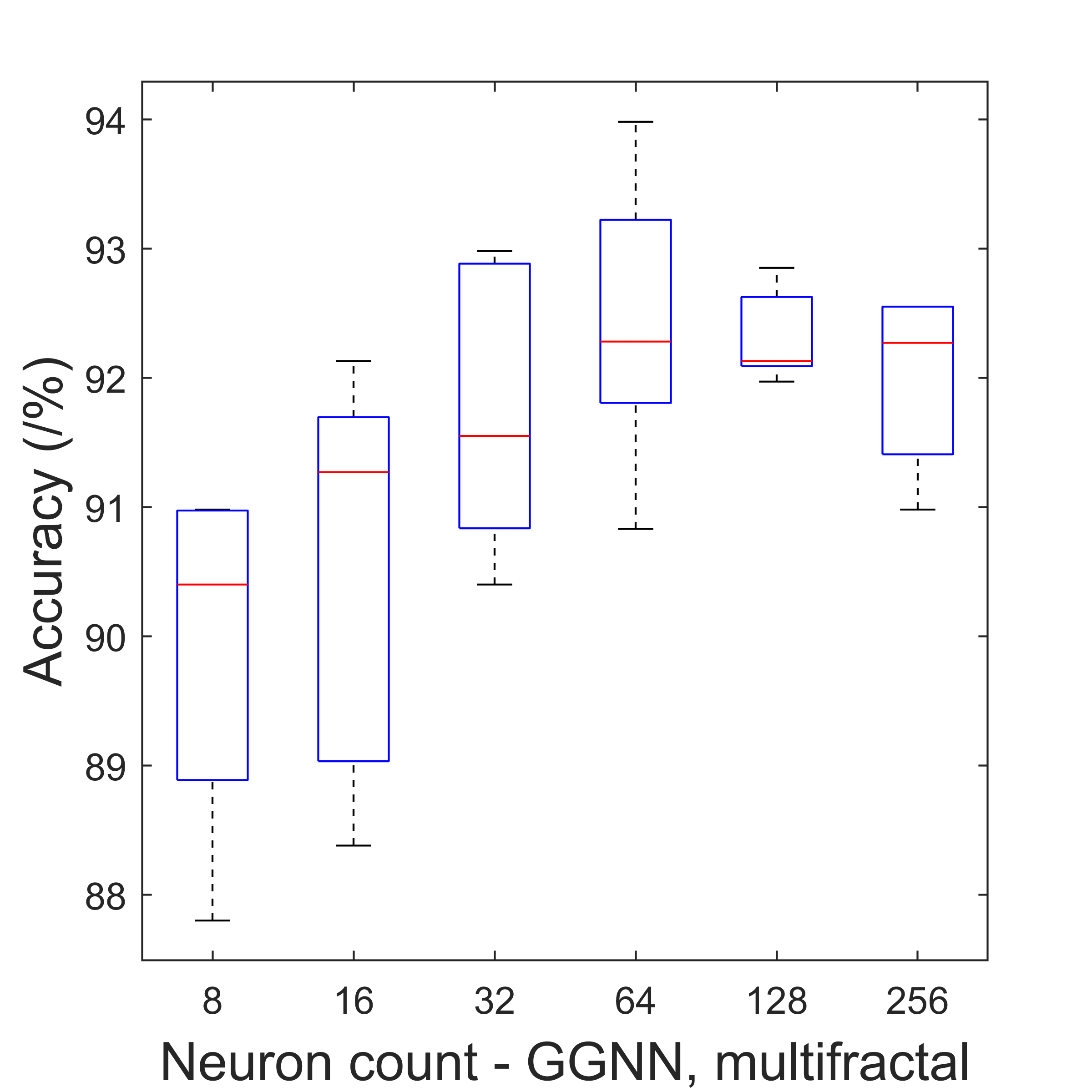}%GGNN
  }
  \vskip -2pt
  \mbox{
    \includegraphics[width=0.31\textwidth,  height=0.31\textwidth]{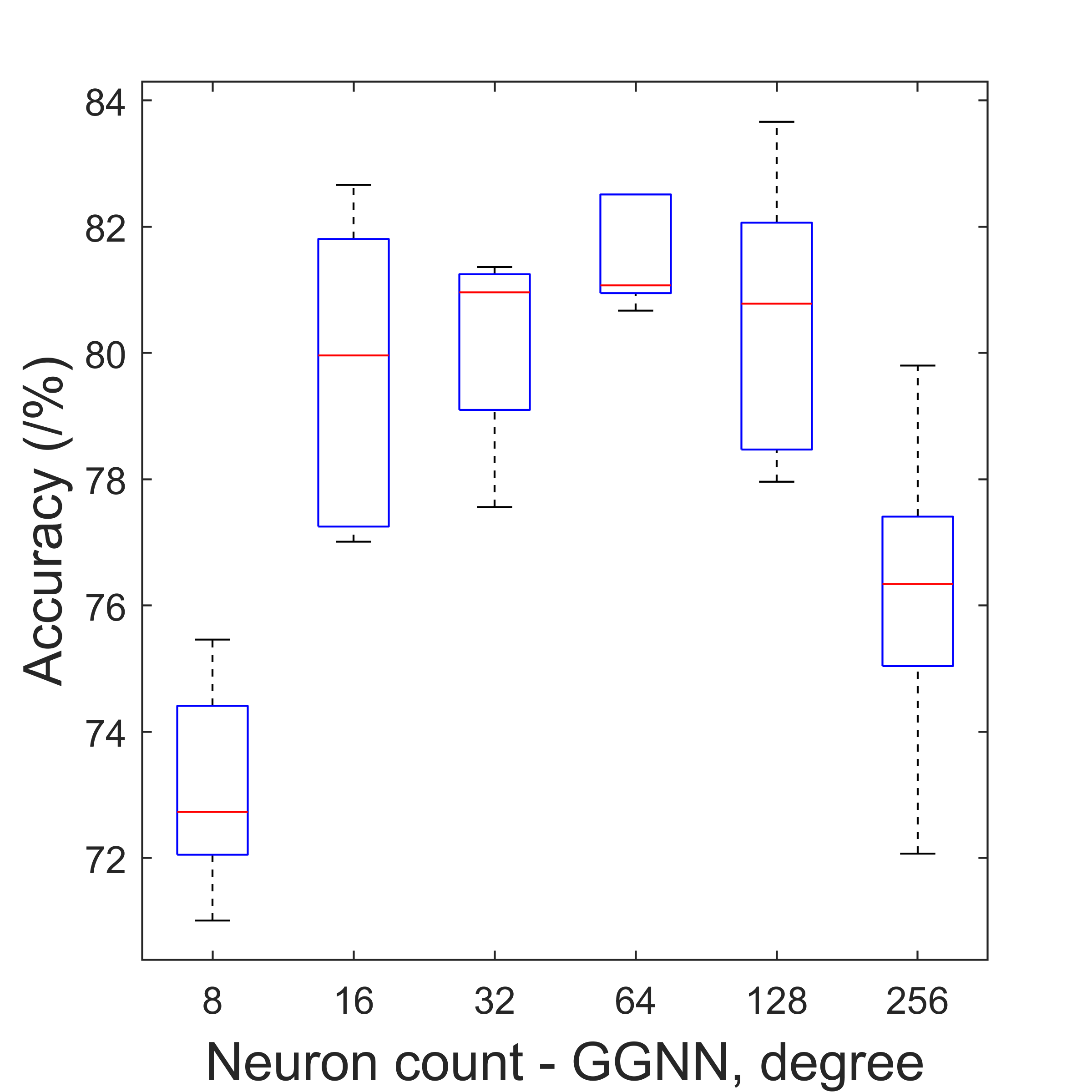}%GCN
    \includegraphics[width=0.31\textwidth,  height=0.31\textwidth]{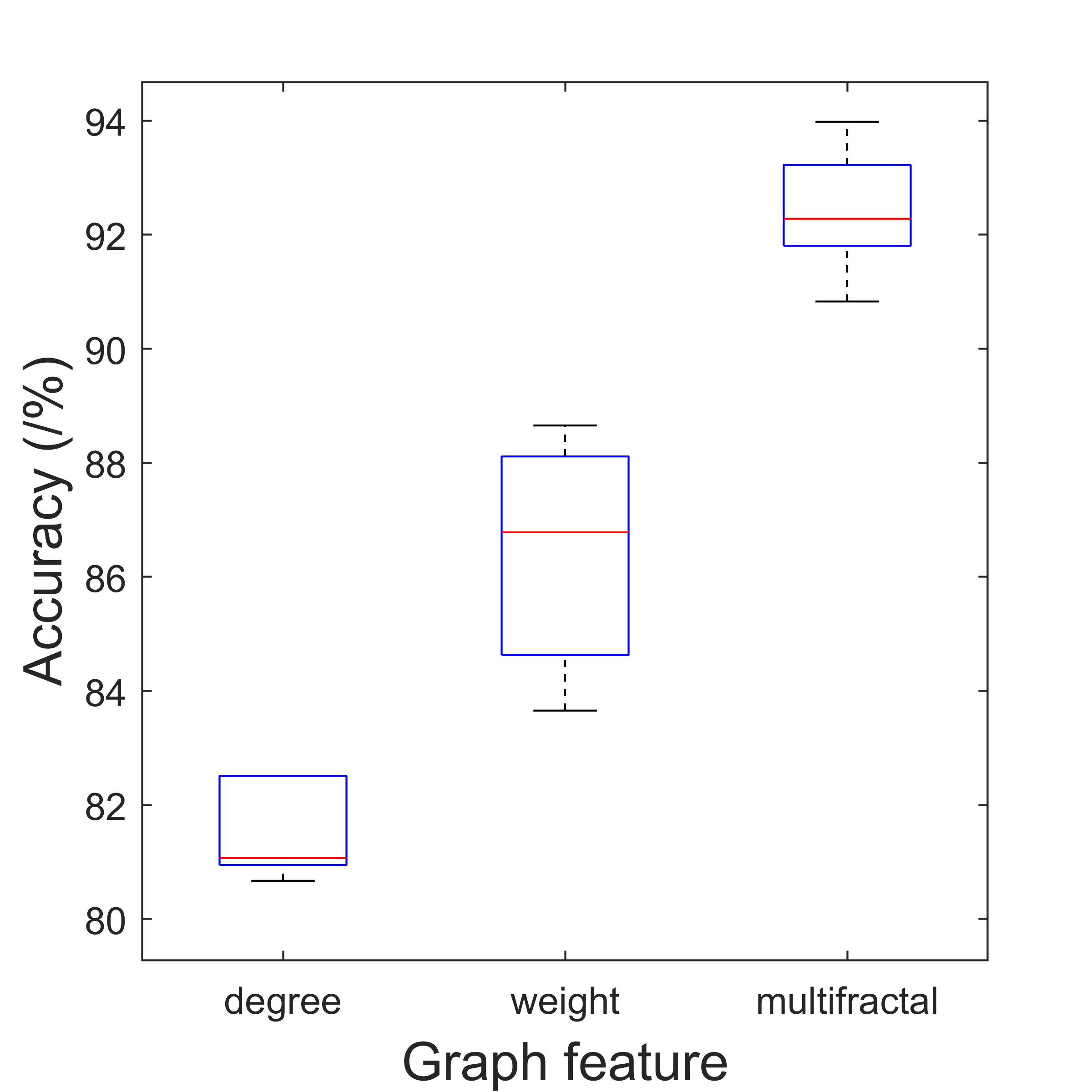}
    \includegraphics[width=0.31\textwidth, height=0.31\textwidth]{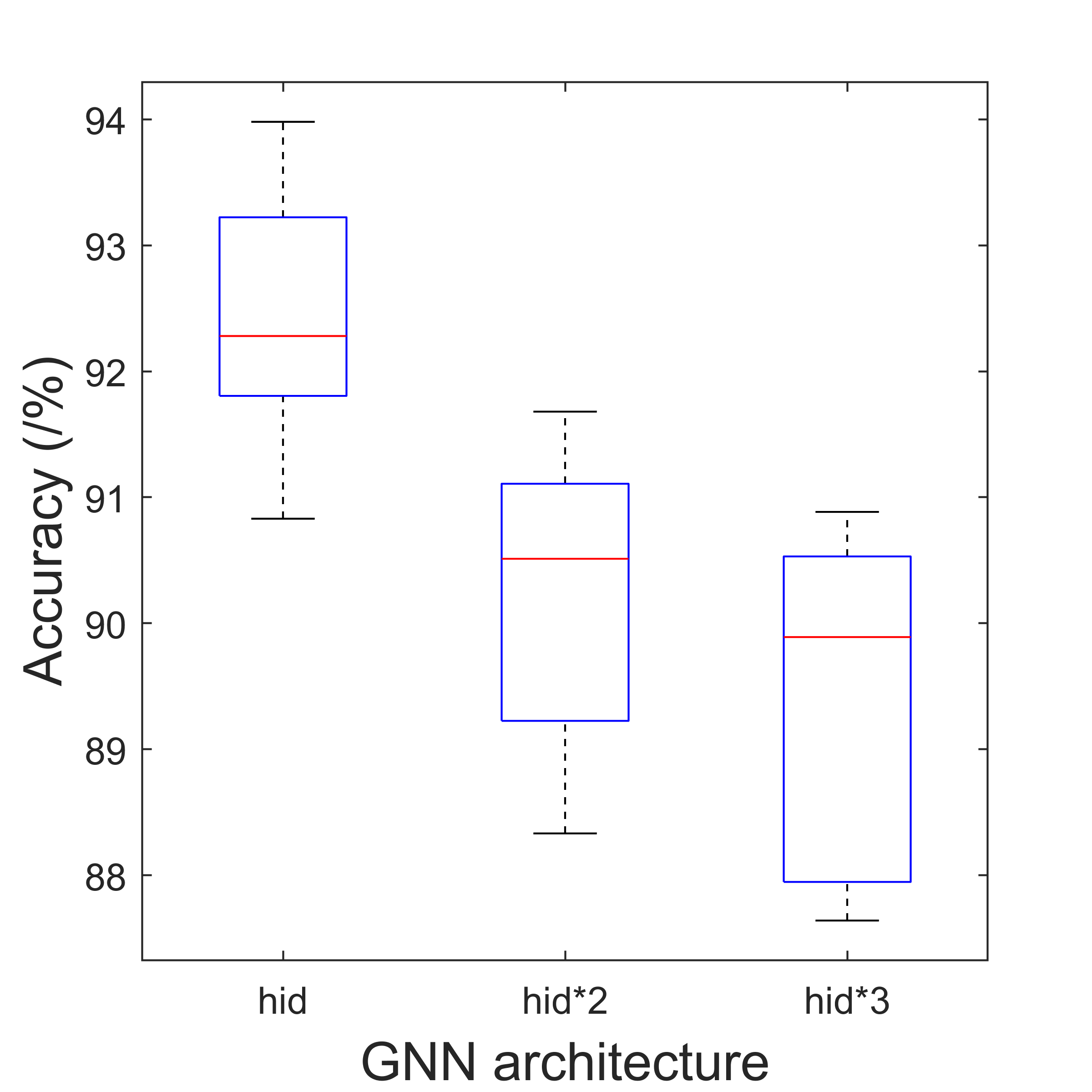}\quad
  }
  \vskip -6pt
  \caption{Ablation study on the impact of various parameters in PGL on the overall accuracy: the number of random walkers, the cut-off range, GNN model, the number of neurons in a hidden layer, different graph features, and GNN architecture.}
\vskip -10pt
\label{fig-ablation}
\end{figure*}

\subsection{Graph Representation Comparison}%Baseline Comparisons}
\begin{wraptable}{r}{0.5\textwidth}
%\parbox{.5\linewidth}{
\small
\caption{Comparison of graph representations.%We compare the graph representation and models used in PGL to the state-of-the-art ProGraML graph representation and the DeepTune and NCC frameworks.
} 
\label{tab-framework}
\scalebox{0.90}{

\begin{tabular}{c|c|c|c|c}
\toprule
%\hline
\textbf{}   & \textbf{Accuracy} & \textbf{Precision} & \textbf{Recall}    & \textbf{$F_1$}         \\ \hline
DeepTune   & 67.31\% $\pm$ 3.89\%    & 0.72      & 0.71 & 0.71          \\ \hline
NCC        & 76.57\% $\pm$ 3.13\%    & 0.80      & 0.80          & 0.80 \\ \hline
ProGraML-GGNN & 80.83\% $\pm$ 3.37\%    & 0.85      & 0.85          & 0.85  \\ \hline
PGL-GCN%-2-32     
& 86.38\% $\pm$ 2.78\%    & 0.90      & 0.90 & 0.90          \\ \hline
PGL-GAT%-2-64     
& 89.65\% $\pm$ 2.24\%    & 0.92      & 0.92 & 0.92          \\ \hline
PGL-GGNN%-1-32     
& 91.39\% $\pm$ 2.59\%    & 0.94      & 0.94 & 0.94          \\ %\hline
\bottomrule
\end{tabular}
}
%}
\vskip -8pt
\end{wraptable}
In order to validate the effectiveness of PGL, we compare it with state-of-the-art techniques in terms of the accuracy of the prediction results on the same dataset \cite{cummins2017end}. We compare against the DeepTune and NCC using the code released by their authors. We also compare our graph representation against the ProGraML graph representation by extracting ProGraML graphs from the C versions of the kernels and training a GGNN on the graphs. As we can see from Table \ref{tab-framework}, DeepTune and NCC can only give less than 80\% accuracy, whereas the ProGraML representation provides 80.83\% accuracy. However, according to neural architecture search, PGL can provide up to 91.39\% accuracy. The reason is that our graph representation of a program differs from the same program with different number of iterations in \textit{for} loops. This enables the graph representation with more information.

\subsection{Ablation Study} \label{sec-ablation}

In this section, we measure different accuracy results from different parameters in PGL by running each experiment 5 times for different number of neurons in the hidden layer, as shown in Figure \ref{fig-ablation} The default parameters include 16 random walkers, cut-off range in multifractal analysis to be the diameter of the graph, GGNN graph model, 64 neurons in a hidden layer, and its architecture to be one input, hidden, and output layer. In the experiments, we vary one parameter while fixing the rest. In the end, we compare the range and distribution of the accuracy for each parameter. 
\paragraph{Random Walkers.} 
We first vary the number of random walkers from 2 to 64 to show how much contribution in random walks. As we can see, increasing the number has a diminishing return beyond 16 walkers as the median accuracy starts at around 73\% at 2 walkers and reaches over 91\% at 16 walkers. It is because when the number of walkers becomes large, some walkers may visit the same nodes, which leads to the same features in multifractal analysis.
\paragraph{Cut-off Range in Multifractal Analysis.}
Next, we vary the cut-off range in multifractal analysis to illustrate how important is multi-fractal analysis. Cut-off range is used in the Dijkstra algorithm in multifractal analysis and defined as the length (sum of edge weights) at which the search is stopped. Therefore, controlling the range allows us to exploit the local structures around a node. In the experiment, we vary it from 2 to 64. As we can see, when the cut-off range is only 2, the accuracy is only 55.97\%. It is because the multifractal analysis in this case does not provide meaningful features to the GNN model. However, as we increase the cut-off range to 64, which is the upper bound of network diameters in the dataset, the accuracy reaches to over 90\% because the multifractal analysis has the full visibility of the graphs and is able to find the correct features.

\paragraph{GNN Model.} PGL provides common interfaces to connect GNNs to the rest of the pipeline. It is flexible enough to support different GNN models. Therefore, in this experiment, we choose three commonly used models to be analyzed, namely, GCN, GAT, and GGNN. As we can see, GGNN provides the highest accuracy with smallest standard deviation (on average over 92\%) whereas GCN and GAT can only provide 81.56\% and 86.4\% on average, respectively. This is mainly due to the fact that GGNN uses the gated recurrent unit (GRU) for long-term
propagation of information across a graph structure \cite{li2015gated}, which enables it to better capture long-range dependencies from the code graphs compared to GCN and GAT. 

\paragraph{Neuron Count.} We validate different combinations of parameters with respect to the number of neurons in a hidden layer on the final accuracy, i.e., (GCN, multifractal), (GAT, multifractal), (GGNN, multifractal), and (GGNN, degree). In general, 64 or 128 neurons provide higher accuracy compared to others. Especially in the case of (GGNN, multifractal), 128 neurons provide smaller standard deviation compared to others. We believe it is because that when using a too small or too large number of neurons, the models cannot accurately learn the hidden structures of a graph.

\paragraph{Graph Feature Embedding.}

\begin{wraptable}{r}{0.5\textwidth}
\small
\caption{Comparison of feature extraction algorithms.%Node feature algorithm compared with the simple node features such as degree and weight, and a state-of-the-art learnable representation of code semantics inst2vec, using the GCN architecture.
}
\scalebox{0.9}{
\begin{tabular}{c|c|c|c|c}

\toprule
%\hline
\textbf{Node Feature}   & \textbf{Accuracy} & \textbf{Precision} & \textbf{Recall}    & \textbf{$F_1$}         \\ \hline
Degree   & 48.54\% $\pm$ 4.33\%    & 0.51      & 0.51 & 0.53          \\ \hline
Weight        & 68.93\% $\pm$ 3.32\%   & 0.73      & 0.73          & 0.73 \\ \hline
inst2vec        & 75.7\% $\pm$ 3.51\%    & 0.76      & 0.77          & 0.78 \\ \hline
PGL     & 91.23\% $\pm$ 2.75\%    & 0.95      & 0.95          & 0.95  \\ %\hline
\bottomrule
\end{tabular}
}
\label{tab-prog}
\end{wraptable}

Next, we vary the graph features from node degree and weights to multifractal properties, using the default GGNN architecture. As we can see from Figure \ref{fig-ablation}, multifractal features can provide at most 93.98\% accuracy and over 90\% on average whereas degree and weight features can only achieve at most 82.51\% and 88.66\% accuracy, respectively. This validates that our proposed graph feature extraction algorithm mentioned in Algorithm 1 can exploit the topological structures of a graph and find the local information around nodes.

In addition, we compare the proposed feature extraction algorithm based on random walk and multifractal analysis concepts rather than simply using the node degree, or edge weight as a feature, and the state-of-the-art \emph{inst2vec} \cite{ben2018neural} on the same GCN architecture to validate the effectiveness of the proposed algorithm. As we can see in Table \ref{tab-prog}, simple node features such as degree and edge weight cannot guarantee stable prediction results on the testing graph data as it only provides up to 72.2\% accuracy. Compared with the state-of-the-art learnable representation of code semantics \emph{inst2vec}, our feature extraction strategy can provide 14.77\% higher accuracy due to the fact that the trained representation of \emph{inst2vec} puts large weights on semantics rather than the code structure. Therefore, our algorithm can achieve better results by quantifying local structures of the code.

\paragraph{GNN Architecture.} 
\begin{wrapfigure}{r}{0.5\textwidth}
\vspace{-5pt}
  \centering
\includegraphics[width=0.46\textwidth, height=0.3\textwidth]{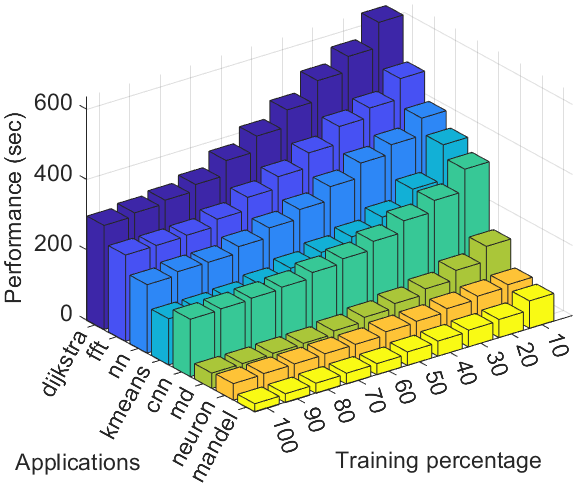}%0.4
  \vskip -7pt
  \caption{The impact of training on the performance of different applications. %X-axis: the percentage of training, Y-axis: the performance measured by execution time in seconds.
  }
\vskip -30pt
\label{figure-app}
\end{wrapfigure}
Finally, in order to see the impact of \emph{deep} GNNs on the final accuracy, we vary the number of hidden layers from 1 to 3. We observe that the average accuracy is decreasing (92.43\%, 90.19\%, and 89.37\%) and the standard deviation is increasing (1.15\%, 1.29\%, and 1.44\%) when the number of hidden layers increases. It is mainly due to the over-squashing issue that tends to occur when increasing the number of layers in GNNs. This causes the information on graphs is compressed and fails to learn long-range signals \cite{alon2020bottleneck}.

\subsection{Parameter Tuning}
As we can see from the ablation study, many hyperparameters could have a significant impact on the overall accuracy of the model we are training. Therefore, in order to find optimal values for the parameters of the GNN model which is later used in the application-level evaluation in Section \ref{sec-appEval}, we rely on grid search to automate parameter tuning. For each parameter discussed in Section \ref{sec-ablation}, we select a range of values to search for and use \textit{GridSearchCV} to improve the accuracy of our model. Experimental results suggest that the number of random walkers be 16, cut-off range be 128, the GNN model be GGNN, the number of neurons be 64, the graph feature be multifractal properties, and the number of hidden layers be 1. 

\subsection{Application-level Evaluation} \label{sec-appEval}
\begin{wrapfigure}{r}{0.5\textwidth}
\vspace{-5pt}
  \centering
\includegraphics[width=0.46\textwidth, height=0.22\textwidth]{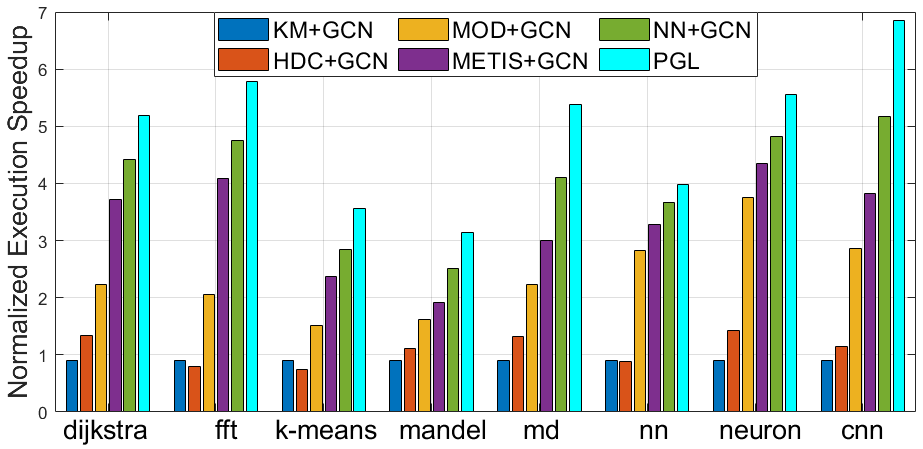}%0.4
  \vskip -7pt
  \caption{Comparison of different partitioning algorithms. %We compare the graph partitioning GAE with different traditional algorithms including K-means, hierarchical divisive clustering, modularity-based community detection, METIS, feed-forward neural network, and GAE, and measure the normalized application execution speedup to validate the benefits of the GAE.
  }
\vskip -20pt
\label{figure-partition}
\end{wrapfigure}

\paragraph{Training.}
First, in order to see how the training step has an impact on the overall accuracy, for each application, we vary the number of epochs used during training of the GGNN model and directly use the model to predict labels for clusters generated from GAE. Finally, we measure the application performance executed in a heterogeneous platform in Table \ref{exp-conf} to figure out how well the model is trained. As we can see from Figure \ref{figure-app}, compared to randomly selecting a label for each cluster (slow execution at 10\%-30\% training), a fully trained model can provide up to 3.8x performance improvement.
\paragraph{Graph Partitioning.}
\begin{wrapfigure}{r}{0.5\textwidth}
\vspace{-5pt}
  \centering
\includegraphics[width=0.46\textwidth, height=0.22\textwidth]{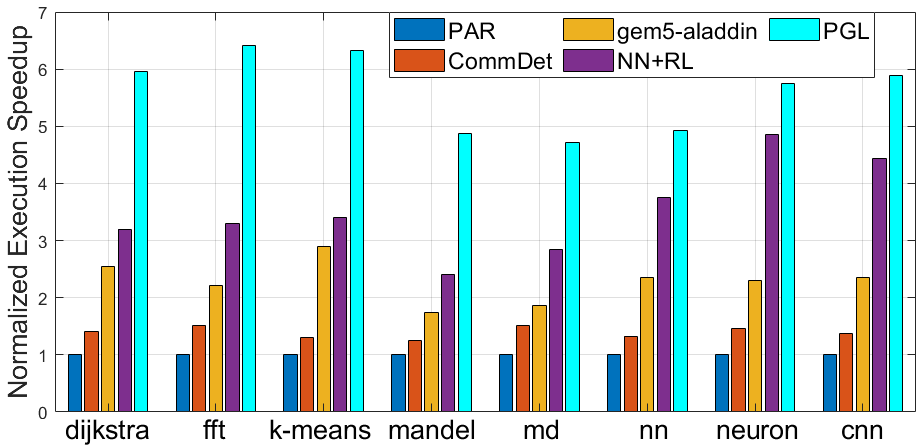}%0.4
  \vskip -7pt
  \caption{Comparison of different frameworks.%The PGL framework. We compare the framework with the traditional thread based parallel programming, modularity based community detection, neural network with reinforcement learning based mapping, and gem5-aladdin in terms of the application performance. We conclude that our approach can achieve 2.02x better compared to the state-of-the-art techniques.
  }
\vskip -15pt
\label{figure-framework}
\end{wrapfigure}
Next, in order to validate the advantages of the GAE used to partition the large input application into small kernels in the PGL framework, we fix the graph neural network as GCN with two hidden layers and 32 neurons per layer, which is used to predict the correct label for each kernel. We compare the GAE with different partitioning algorithms such as K-means (KM), hierarchical divisive clustering (HDC), modularity-based community detection (MOD), METIS, and feed-forward neural network (NN) in terms of the total application execution speedup. As shown in Figure \ref{figure-partition}, for the partitioning models without machine learning such as KM, HDC, MOD, and METIS, the normalized execution speedup is smaller compared to the learning models such as NN and GAE. It is mainly because the kernels after graph partitioning are not well recognized by the GCN model. For the learning models, GAE outperforms NN by up to 32\% in a sense that the GAE takes into account the graph structures of code. 

\paragraph{The PGL Framework.}
The proposed PGL framework is able to predict which code segments run best on a specific processor. Therefore, in order to validate the framework including the GAE and GNN models, we use the trained models discussed in Section \ref{sec:framework} to predict each application in Table \ref{table-data}. As shown in Figure \ref{figure-framework}, we use the traditional thread based parallel programming running on CPUs as our baseline and compare the PGL framework with community detection, neural network with reinforcement learning, and gem5-aladdin. We observe that the PGL framework can provide up to 6.42x speedup compared to the baseline and 2.02x speedup higher compared to the state-of-the-art.

%% file: sec-conclusion.tex
\section{Conclusion}
\label{sec:conc}

We proposed an end-to-end learnable PGL framework to predict which code segments run best on a specific hardware device. We first develop a node feature extraction algorithm based on random walks and multifractal analysis concepts to quantify the local structures of a program. Next, we build the GAE together with a decoder and spectral clustering to find cluster partition from the distance matrix. Then, we use graph neural networks as the learning model to predict the type of each cluster. Our evaluation on 32 CPUs and 32 GPUs concludes that the PGL framework can provide up to 6.42x speedup compared to the baseline and 2.02x higher speedup compared to the state-of-the-art technique.

%% file: sec-appendix.tex
\newpage
\appendix
\onecolumn
\section*{Appendix}
\section{Random Walk based Fractal Analysis}

\begin{algorithm}

\small
    \caption{Random Walk based Fractal Analysis}
  \begin{algorithmic}[1]
  \label{algo-random}
    \STATE \textbf{INPUTS}: An LLVM graph $G$ with $N$ nodes, $K$ random walks, $D$ walk length, $Q$ distortion factors 
    \STATE \textbf{OUTPUT}: $N$ by ($K\times Q$) node features $F$
    \STATE Create a feature matrix $F$ of size $N$ by $K\times Q$
    \FOR{each node indexed by $i$ in the graph $G$}
        \FOR{K times}
        \STATE /* Perform a random walk */
        \FOR{D times}
            \STATE Calculate the probability of the transition to node $j$
            \STATE Select the next node and set it as the current node
        %\UNTIL{$D$ times}
        \ENDFOR
        \STATE /* So far we find the destination node denoted as $j$ */
        \STATE /* Find the subgraph $SG$ starting from $i$ to $j$ */
        \STATE Backtrack the node $j$ to find all nodes until the node $i$
        \STATE /* Find the generalized fractal dimension from the $SG$ */
        \STATE Set the distortion factor $q$ to be a vector of -10 to 10
        \STATE $w$ = GFD($SG$, $q$)
        \STATE $F$[$i$].append($w$)
        %\UNTIL{$K$ times
        \ENDFOR
        
    \ENDFOR
%  \end{algorithmic}
%\end{algorithm}
% \algrule
\\\hrulefill 
%\begin{algorithm}
%\small
%    \caption{Generalized Fractal Dimension from Multifractal Analysis}
%  \begin{algorithmic}[1]  
    % \Procedure{GFD}{$G$}
    \STATE \textbf{Function}: GFD
    \STATE \textbf{INPUTS}: a graph $G$, distortion factor $q$ of size $Q$
    \STATE \textbf{OUTPUT}: generalized fractal dimension of size $Q$
    \STATE diameter = Diam($G$) 
    \FOR{each node $i$ in the graph $G$}
        \STATE Calculate the shortest path length from $i$ to every node
        \STATE Calculate the ratio of nodes to be covered with a box size $l$
    \ENDFOR
    \FOR{each $qv$ in $q$}
        %\STATE Calculate $Zq$ using Eq. (X)
        \STATE Apply linear regression to find the exponent $\tau$ in Eq. (2)
        \STATE tau.append($\tau$)
    \ENDFOR
    \STATE gfd = tau / ($q$ - 1)
    % \EndProcedure
  \end{algorithmic}
\end{algorithm}

\section{GAE Partitioning}

\begin{algorithm}
\small
    \caption{GAE Partitioning}
  \begin{algorithmic}[1]
  \label{algo-partition}
    \STATE \textbf{INPUTS}: A graph $G$ and a feature matrix $X$
    \STATE \textbf{OUTPUT}: A cluster partition
    \REPEAT
    \STATE Perform the GAE with two graph convolutional layers to get the embedding $Z$
    \STATE Calculate the symmetric distance matrix $\mathbf{D}$ by $\hat{\mathbf{A}}=ZZ^T, \mathbf{D}=\frac{1}{2}(|\hat{\mathbf{A}}|+|\hat{\mathbf{A}}|^{\top})$
    \STATE Obtain the partition via spectral clustering on $\mathbf{D}$
    \UNTIL{99\% of nodes in the partition are stabilized.}

  \end{algorithmic}
\end{algorithm}

Given the graph $G=(V,E)$ with an adjacency matrix $\mathbf{A}$ and node features in an $N\times D$ matrix $\mathbf{X}$, we apply the graph auto-encoder (GAE) model introduced in \cite{kipf2016variational} with two graph convolutional layers. We calculate embeddings $\mathbf{Z}$ and the reconstructed matrix $\hat{\mathbf{A}}$ as follows:
\begin{align}
    \hat{\mathbf{A}}=\sigma(\mathbf{ZZ}^{\top}), \text{with   } \mathbf{Z}=GCN(\mathbf{X,A})
\end{align}

After we obtain the node embeddings via GAE, we use spectral clustering \cite{von2007tutorial} on the node embeddings for the graph partitioning. The overall workflow of this stage is shown in Algorithm 2%, mentioned in the supplementary material
. Specifically, we first perform the GAE with two graph convolutional layers to learn the latent embedding $\mathbf{Z}$. Next, we maintain an inner product decoder $\hat{\mathbf{A}}=\mathbf{ZZ}^{\top}$ to learn the pairwise distance between nodes. We then perform spectral clustering after calculating the symmetric and non-negative distance matrix $\mathbf{D}=\frac{1}{2}(|\hat{\mathbf{A}}|+|\hat{\mathbf{A}}|^{\top})$.

\subsection{Graph Neural Networks}

\paragraph{Graph Convolutional Network.}
We consider a multi-layer graph convolutional network (GCN) with the layer-wise propagation rule proposed in \cite{kipf2016semi}. Assume ${\mathbf{H}}^{(l)}\in \mathbb{R}^{n \times d}$ is the matrix of activations in the $l$-th layer, according to the propagation rule, we have 
\begin{equation}
    \mathbf{H}^{(l+1)} = \sigma(\mathbf{\tilde{D}}^{-\frac{1}{2}}\mathbf{\tilde{A}}\mathbf{\tilde{D}}^{-\frac{1}{2}}\mathbf{H}^{(l)}\mathbf{\Theta}^{(l)})
\end{equation}
where $\mathbf{\tilde{A}} = \mathbf{A} + \mathbf{I}_{n}$ is the weight matrix of the graph, $\mathbf{I}_{n}$ is an identity matrix, $\mathbf{D} \in \mathbb{R}^{n \times n}$ is the diagonal degree matrix of the graph, and $\sigma(\cdot)$ denotes the activation function. $\mathbf{\Theta}$ is a layer-specific trainable weight matrix. $\mathbf{H}^{(0)} = \mathbf{X}$ is the input feature matrix of the graph.

\paragraph{Graph Attention Network.}
Unlike the GCNs discussed above, where the node neighborhoods are aggregated with equal weights, graph attention network (GAT) introduces an attention mechanism into GCNs to allow variance in the influences of neighbors. We use the GAT introduced in~\cite{velivckovic2017graph,lee2019attention} and define the  propagation rule in GAT as:

\begin{equation}
    \mathbf{h}_i^{(l+1)} = \sigma(\sum_{j\in {\tilde{\mathcal{N}}(i)}}{\alpha}_{ij}^{(l)}{\mathbf{h}_j^{(l)}}\mathbf{\Theta}^{(l)})
\end{equation}
where ${\alpha}_{ij}^{(l)}$ is node $v_i$'s attention to node $v_j$ in the $l$-th layer:

\begin{equation}
    {\alpha}_{ij}^{(l)}=\frac{\text{exp}(\text{LeakyReLU}(\mathcal{F}(\mathbf{h}_i^{(l)}\mathbf{\Theta}^{(l)},\mathbf{h}_j^{(l)}\mathbf{\Theta}^{(l)})))}{\sum_{k\in {\tilde{\mathcal{N}}(i)}}\text{exp}(\text{LeakyReLU}(\mathcal{F}(\mathbf{h}_i^{(l)}\mathbf{\Theta}^{(l)},\mathbf{h}_k^{(l)}\mathbf{\Theta}^{(l)})))}
\end{equation}
where $\tilde{\mathcal{N}}(i)$ is the set of neighboring nodes of node $i$ in the graph, $\mathcal{F}(\cdot,\cdot)$ is a function to be learned. In our experiments, we use a single-layer feedforward neural network for the attention mechanism parameterized by a weight vector, and apply the LeakyReLU nonlinearity. 

\paragraph{Gated Graph Neural Network.}
Gated graph neural network (GGNN) incorporates gate mechanism like Gate Recurrent Units (GRU) \cite{cho2014learning} or Long Short-term Memory (LSTM) \cite{hochreiter1997long} in the propagation stage in the GNN models to improve the long-term information propagation across the graph. In this work, we use the GGNN with GRU introduced in \cite{li2015gated} and define the recurrence of the propagation as follows:
\begin{align*}
    & \mathbf{a}_i^t = \mathbf{A}_i^T[\mathbf{h}_1^{t-1}\cdots \mathbf{h}_N^{t-1}]^T + \mathbf{b} \\
    & \mathbf{z}_i^t = \sigma (\mathbf{W}^z\mathbf{a}_i^t + \mathbf{U}^z\mathbf{h}_i^{t-1}) \\
    & \mathbf{r}_i^t = \sigma (\mathbf{W}^r\mathbf{a}_i^t + \mathbf{U}^r\mathbf{h}_i^{t-1})  \\  
    & \tilde{\mathbf{h}_i^t} = \tanh(\mathbf{W}\mathbf{a}_i^t + \mathbf{U}(\mathbf{r}_i^t\odot \mathbf{h}_i^{t-1}))\\
    & \mathbf{h}_i^t = (1 - \mathbf{z}_i^t)\odot \mathbf{h}_i^{t-1} + \mathbf{z}_i^t \odot \tilde{\mathbf{h}_i^t}
\end{align*}

where the node $i$ first aggregates message from its neighbors, and $\mathbf{A}_i$ is the sub-matrix of the graph adjacency matrix $A$, which denotes the connections between node $i$ and its neighbors. $\mathbf{z}$ and $\mathbf{r}$ are the update and reset gates, respectively. As such, the GRU update functions incorporate information from the other nodes and from the previous timestamp to update the hidden state of each node $i$.

%\textcolor{blue}{add some descriptions for the architecture of these GNNs for the prediction task. }
%e.g. in our experiment, we use these GNNs with fully connected layer for the classification task. Specifically, for GCN, we use 2 GC layers and 1...For GAE
Once the feature embedding is learned from the GNN models, we use two fully connected feed-forward neural network layers to predict the correct label for each kernel. %In the evaluation, we vary the number of layers used in the GNN models and the number of neurons per layer to measure the total accuracy.

\section{Experiment Setup and Benchmarks}
\subsection{Experiment Setup}
\begin{table}[H]
%\centering
%\vskip -15pt
%\scalebox{1}{
%\scriptsize
%\parbox{1\linewidth}{
\centering
\caption{Configuration parameters}
\begin{tabular}{ l|l|l }
\multicolumn{3}{l}{l} \\ \toprule
\multirow{3}{*}{\textbf{CPU}} & Cores & 32 cores, 16 MSHRs \\ \cline{2-3}
 & Clock frequency & 2.4 GHz\\ \cline{2-3}
 & L1 private cache & 64KB, 4-way associative \\
 && 32-byte blocks \\ \cline{2-3}
 & L2 shared cache & 256KB, distributed \\ \cline{2-3}
 & Memory & 4 GB, 8 GB/s bandwidth \\ \hline
\multirow{4}{*}{\textbf{GPU}} & Core & 32 \\ \cline{2-3}
 & Clock frequency & 575 MHz\\ \cline{2-3}
 & Memory capacity & 768 MB \\ \cline{2-3}
 & Memory bandwidth & 86.4 GB/s \\ \hline
\multirow{3}{*}{\textbf{Network}} & Topology & Mesh \\ \cline{2-3}
 & Routing algorithm & XY routing \\ \cline{2-3}
 & Flow control & Virtual channel flit-based  \\ %\hline
 \bottomrule
\end{tabular}
%}
\label{exp-conf}
\end{table}
%}

\subsection{Benchmarks}
\begin{table}[H]
%\hspace{9pt}
\centering
\caption{Applications and descriptions. We use the following eight benchmarks to validate the benefits of the PGL framework whereas we use the dataset \cite{cummins2017end} to train the graph neural network in the framework.}
%\centering
%\resizebox{\columnwidth}{!}{%
\begin{tabular}{l|l|l} %\hline
\toprule
\textbf{Application}&\textbf{Description}&\textbf{Input Size} \\ \hline
dijkstra&Find the shortest path&100 nodes\\ \hline
fft&Fast Fourier transform&vector of size 4096\\ \hline
kmeans&K cluster partitioning&256 2D tuples\\ \hline
mandel&Calculate Mandelbrot set&4092 points\\ \hline
md&Molecular dynamics&1024 particles\\ \hline
nn&Neural network&5 hidden FC layers\\ \hline
neuron&ReLU neurons&1024 neurons\\ \hline
cnn&Conv. neural network&conv-pool-FC\\ %\hline
\bottomrule
\end{tabular}
\vskip -10pt
\label{table-data}
%\vskip -10pt
%}
\end{table}

\begin{table}[H]

\centering
\small
\caption{Application graph statistics.}
\vskip -5pt
\begin{tabular}{l|l|l|l}
\toprule
%\hline
\textbf{Application}        & \textbf{No. Nodes}   & \textbf{No. Edges} & \textbf{Avg Path Length}\\ \hline
dijkstra      & 502,897   & 588,046  & 17.36\\ \hline
fft           & 456,183   & 572,053 & 15.54\\ \hline
kmeans       & 705,184   & 839,125 & 22.18\\ \hline
mandel        & 235,051   & 260,042 & 11.67\\ \hline
md            & 1,799,353 & 2,361,213 & 34.29\\ \hline
nn            & 227,766   & 286,714 & 19.22\\ \hline
neuron        & 987,184   & 1174,843 & 52.75\\ \hline
cnn           & 361,464   & 520,596 & 13.34\\ %\hline
\bottomrule
\end{tabular}

\label{table-sta}
\vskip -10pt
\end{table}

\section{Run-time Mapping}
At run-time, the tasks generated from each application are mapped onto the system. The mapping exploits and optimizes the parallelism while considering data communication between tasks and resource utilization \cite{ma2021distributed}.

In order to improve performance, the run-time mapping should exploit and optimize the parallelism while considering data communication between tasks and resource utilization. Therefore, the run-time mapping algorithm takes as inputs the tasks, their interactions, and data communication and schedules a mapping from tasks to processors with the objective of improving application performance. If data are transferred between two different tasks, then the greedy run-time scheduler tried to allocate two available cores that are the closest based on the Manhattan distance.

%% file: paper.bbl
\begin{thebibliography}{10}

\bibitem{krishnan2019air}
Srivatsan Krishnan, Behzad Borojerdian, William Fu, Aleksandra Faust, and
  Vijay~Janapa Reddi.
\newblock Air learning: An ai research platform for algorithm-hardware
  benchmarking of autonomous aerial robots.
\newblock {\em arXiv preprint arXiv:1906.00421}, 2019.

\bibitem{haj2019view}
Ameer Haj-Ali, Nesreen~K Ahmed, Ted Willke, Joseph Gonzalez, Krste Asanovic,
  and Ion Stoica.
\newblock A view on deep reinforcement learning in system optimization.
\newblock {\em arXiv preprint arXiv:1908.01275}, 2019.

\bibitem{xiao2019self}
Yao Xiao, Shahin Nazarian, and Paul Bogdan.
\newblock Self-optimizing and self-programming computing systems: A combined
  compiler, complex networks, and machine learning approach.
\newblock {\em IEEE Transactions on Very Large Scale Integration (VLSI)
  Systems}, 27(6):1416--1427, 2019.

\bibitem{xiao2017load}
Yao Xiao, Yuankun Xue, Shahin Nazarian, and Paul Bogdan.
\newblock A load balancing inspired optimization framework for exascale
  multicore systems: A complex networks approach.
\newblock In {\em 2017 IEEE/ACM International Conference on Computer-Aided
  Design (ICCAD)}, pages 217--224. IEEE, 2017.

\bibitem{cummins2017end}
Chris Cummins, Pavlos Petoumenos, Zheng Wang, and Hugh Leather.
\newblock End-to-end deep learning of optimization heuristics.
\newblock In {\em 2017 26th International Conference on Parallel Architectures
  and Compilation Techniques (PACT)}, pages 219--232. IEEE, 2017.

\bibitem{mirhoseini2017device}
Azalia Mirhoseini, Hieu Pham, Quoc~V Le, Benoit Steiner, Rasmus Larsen, Yuefeng
  Zhou, Naveen Kumar, Mohammad Norouzi, Samy Bengio, and Jeff Dean.
\newblock Device placement optimization with reinforcement learning.
\newblock In {\em International Conference on Machine Learning}, pages
  2430--2439. PMLR, 2017.

\bibitem{cummins2020programl}
Chris Cummins, Zacharias Fisches, Tal Ben-Nun, Torsten Hoefler, Michael
  O'Boyle, and Hugh Leather.
\newblock {ProGraML: A Graph-based Program Representation for Data Flow
  Analysis and Compiler Optimizations}.
\newblock In {\em International Conference on Machine Learning (ICML)}, 2021.

\bibitem{grewe2013portable}
Dominik Grewe, Zheng Wang, and Michael~FP O'Boyle.
\newblock Portable mapping of data parallel programs to opencl for
  heterogeneous systems.
\newblock In {\em Proceedings of the 2013 IEEE/ACM International Symposium on
  Code Generation and Optimization (CGO)}, pages 1--10. IEEE, 2013.

\bibitem{xue2017reliable}
Yuankun Xue and Paul Bogdan.
\newblock Reliable multi-fractal characterization of weighted complex networks:
  algorithms and implications.
\newblock {\em Scientific reports}, 7(1):1--22, 2017.

\bibitem{ashouri2018survey}
Amir~H Ashouri, William Killian, John Cavazos, Gianluca Palermo, and Cristina
  Silvano.
\newblock A survey on compiler autotuning using machine learning.
\newblock {\em ACM Computing Surveys (CSUR)}, 51(5):1--42, 2018.

\bibitem{li2020deep}
Mingzhen Li, Yi~Liu, Xiaoyan Liu, Qingxiao Sun, Xin You, Hailong Yang, Zhongzhi
  Luan, Lin Gan, Guangwen Yang, and Depei Qian.
\newblock The deep learning compiler: A comprehensive survey.
\newblock {\em IEEE Transactions on Parallel and Distributed Systems},
  32(3):708--727, 2020.

\bibitem{zhou2020transferable}
Yanqi Zhou, Sudip Roy, Amirali Abdolrashidi, Daniel Wong, Peter Ma, Qiumin Xu,
  Hanxiao Liu, Mangpo~Phitchaya Phothilimtha, Shen Wang, Anna Goldie, et~al.
\newblock Transferable graph optimizers for ml compilers.
\newblock {\em arXiv preprint arXiv:2010.12438}, 2020.

\bibitem{haj2020neurovectorizer}
Ameer Haj-Ali, Nesreen~K Ahmed, Ted Willke, Yakun~Sophia Shao, Krste Asanovic,
  and Ion Stoica.
\newblock Neurovectorizer: End-to-end vectorization with deep reinforcement
  learning.
\newblock In {\em Proceedings of the 18th ACM/IEEE International Symposium on
  Code Generation and Optimization}, pages 242--255, 2020.

\bibitem{haj2019learning}
Ameer Haj-Ali, Nesreen~K Ahmed, Ted Willke, Sophia Shao, Krste Asanovic, and
  Ion Stoica.
\newblock Learning to vectorize using deep reinforcement learning.
\newblock In {\em Neurips workshop on Machine Learning for Systems.}, 2019.

\bibitem{jinnai2019knossos}
Yuu Jinnai, Arash Mehrjou, Kamil Ciosek, Anna Mitenkova, Alan Lawrence, Tom
  Ellis, Ryota Tomioka, Simon~Peyton Jones, and Andrew Fitzgibbon.
\newblock Knossos: Compiling ai with ai.
\newblock 2019.

\bibitem{jia2019taso}
Zhihao Jia, Oded Padon, James Thomas, Todd Warszawski, Matei Zaharia, and Alex
  Aiken.
\newblock Taso: optimizing deep learning computation with automatic generation
  of graph substitutions.
\newblock In {\em Proceedings of the 27th ACM Symposium on Operating Systems
  Principles}, pages 47--62, 2019.

\bibitem{jia2018beyond}
Zhihao Jia, Matei Zaharia, and Alex Aiken.
\newblock Beyond data and model parallelism for deep neural networks.
\newblock {\em arXiv preprint arXiv:1807.05358}, 2018.

\bibitem{nguyen2018deep}
Anh~Tuan Nguyen, Trong~Duc Nguyen, Hung~Dang Phan, and Tien~N Nguyen.
\newblock A deep neural network language model with contexts for source code.
\newblock In {\em 2018 IEEE 25th International Conference on Software Analysis,
  Evolution and Reengineering (SANER)}, pages 323--334. IEEE, 2018.

\bibitem{alon2019code2vec}
Uri Alon, Meital Zilberstein, Omer Levy, and Eran Yahav.
\newblock code2vec: Learning distributed representations of code.
\newblock {\em Proceedings of the ACM on Programming Languages}, 3(POPL):1--29,
  2019.

\bibitem{ben2018neural}
Tal Ben-Nun, Alice~Shoshana Jakobovits, and Torsten Hoefler.
\newblock Neural code comprehension: A learnable representation of code
  semantics.
\newblock {\em arXiv preprint arXiv:1806.07336}, 2018.

\bibitem{brauckmann2020compiler}
Alexander Brauckmann, Andr{\'e}s Goens, Sebastian Ertel, and Jeronimo
  Castrillon.
\newblock Compiler-based graph representations for deep learning models of
  code.
\newblock In {\em Proceedings of the 29th International Conference on Compiler
  Construction}, pages 201--211, 2020.

\bibitem{li2019graph}
Yujia Li, Chenjie Gu, Thomas Dullien, Oriol Vinyals, and Pushmeet Kohli.
\newblock Graph matching networks for learning the similarity of graph
  structured objects.
\newblock In {\em International Conference on Machine Learning}, pages
  3835--3845. PMLR, 2019.

\bibitem{mou2016convolutional}
Lili Mou, Ge~Li, Lu~Zhang, Tao Wang, and Zhi Jin.
\newblock Convolutional neural networks over tree structures for programming
  language processing.
\newblock In {\em Proceedings of the AAAI Conference on Artificial
  Intelligence}, volume~30, 2016.

\bibitem{grover2016node2vec}
Aditya Grover and Jure Leskovec.
\newblock node2vec: Scalable feature learning for networks.
\newblock In {\em Proceedings of the 22nd ACM SIGKDD international conference
  on Knowledge discovery and data mining}, pages 855--864, 2016.

\bibitem{wu2020comprehensive}
Zonghan Wu, Shirui Pan, Fengwen Chen, Guodong Long, Chengqi Zhang, and S~Yu
  Philip.
\newblock A comprehensive survey on graph neural networks.
\newblock {\em IEEE transactions on neural networks and learning systems},
  2020.

\bibitem{zhou2018graph}
Jie Zhou, Ganqu Cui, Zhengyan Zhang, Cheng Yang, Zhiyuan Liu, Lifeng Wang,
  Changcheng Li, and Maosong Sun.
\newblock Graph neural networks: A review of methods and applications.
\newblock {\em arXiv preprint arXiv:1812.08434}, 2018.

\bibitem{kipf2016variational}
Thomas~N Kipf and Max Welling.
\newblock Variational graph auto-encoders.
\newblock {\em arXiv preprint arXiv:1611.07308}, 2016.

\bibitem{von2007tutorial}
Ulrike Von~Luxburg.
\newblock A tutorial on spectral clustering.
\newblock {\em Statistics and computing}, 17(4):395--416, 2007.

\bibitem{kipf2016semi}
Thomas~N Kipf and Max Welling.
\newblock Semi-supervised classification with graph convolutional networks.
\newblock {\em arXiv preprint arXiv:1609.02907}, 2016.

\bibitem{velivckovic2017graph}
Petar Veli{\v{c}}kovi{\'c}, Guillem Cucurull, Arantxa Casanova, Adriana Romero,
  Pietro Lio, and Yoshua Bengio.
\newblock Graph attention networks.
\newblock {\em arXiv preprint arXiv:1710.10903}, 2017.

\bibitem{lee2019attention}
John~Boaz Lee, Ryan~A Rossi, Sungchul Kim, Nesreen~K Ahmed, and Eunyee Koh.
\newblock Attention models in graphs: A survey.
\newblock {\em ACM Transactions on Knowledge Discovery from Data (TKDD)},
  13(6):1--25, 2019.

\bibitem{li2015gated}
Yujia Li, Daniel Tarlow, Marc Brockschmidt, and Richard Zemel.
\newblock Gated graph sequence neural networks.
\newblock {\em arXiv preprint arXiv:1511.05493}, 2015.

\bibitem{fortunato2010community}
Santo Fortunato.
\newblock Community detection in graphs.
\newblock {\em Physics reports}, 486(3-5):75--174, 2010.

\bibitem{lasalle2015improving}
Dominique LaSalle, Md~Mostofa~Ali Patwary, Nadathur Satish, Narayanan Sundaram,
  Pradeep Dubey, and George Karypis.
\newblock Improving graph partitioning for modern graphs and architectures.
\newblock In {\em Proceedings of the 5th Workshop on Irregular Applications:
  Architectures and Algorithms}, pages 1--4, 2015.

\bibitem{shao2016co}
Yakun~Sophia Shao, Sam~Likun Xi, Vijayalakshmi Srinivasan, Gu-Yeon Wei, and
  David Brooks.
\newblock Co-designing accelerators and soc interfaces using gem5-aladdin.
\newblock In {\em 2016 49th Annual IEEE/ACM International Symposium on
  Microarchitecture (MICRO)}, pages 1--12. IEEE, 2016.

\bibitem{alon2020bottleneck}
Uri Alon and Eran Yahav.
\newblock On the bottleneck of graph neural networks and its practical
  implications.
\newblock {\em arXiv preprint arXiv:2006.05205}, 2020.

\bibitem{cho2014learning}
Kyunghyun Cho, Bart Van~Merri{\"e}nboer, Caglar Gulcehre, Dzmitry Bahdanau,
  Fethi Bougares, Holger Schwenk, and Yoshua Bengio.
\newblock Learning phrase representations using rnn encoder-decoder for
  statistical machine translation.
\newblock {\em arXiv preprint arXiv:1406.1078}, 2014.

\bibitem{hochreiter1997long}
Sepp Hochreiter and J{\"u}rgen Schmidhuber.
\newblock Long short-term memory.
\newblock {\em Neural computation}, 9(8):1735--1780, 1997.

\bibitem{ma2021distributed}
Guixiang Ma, Yao Xiao, Theodore Willke, Nesreen Ahmed, Shahin Nazarian, and
  Paul Bogdan.
\newblock A distributed graph-theoretic framework for automatic parallelization
  in multi-core systems.
\newblock {\em Proceedings of Machine Learning and Systems}, 3:550--568, 2021.

\end{thebibliography}
